\documentclass{article} % For LaTeX2e
\usepackage{iclr2022_conference,times}

\usepackage{hyperref}
\usepackage{url}

\usepackage{amsmath, amssymb, amsthm, amsbsy}
\usepackage{bm}
\usepackage{graphicx}
\usepackage{xspace}
\usepackage{subcaption}
\usepackage{url}
\usepackage{enumerate}
\usepackage{sidecap}
\usepackage{standalone}
\usepackage{caption}
\usepackage{multirow}
\usepackage{dsfont}
\usepackage[subtle]{savetrees}
\usepackage{booktabs} 
\usepackage{colortbl} 
\usepackage{xfrac}
\usepackage{pgfplots}
\pgfplotsset{compat=1.16}
\usepgfplotslibrary{groupplots}
\usetikzlibrary{patterns}
\PassOptionsToPackage{sort}{natbib}
\usepackage[utf8]{inputenc} % allow utf-8 input
\usepackage[T1]{fontenc}    % use 8-bit T1 fonts
\usepackage{hyperref}       % hyperlinks
\usepackage{url}            % simple URL typesetting
\usepackage{booktabs}       % professional-quality tables
\usepackage{amsfonts}       % blackboard math symbols
\usepackage{amsmath}
\usepackage{amstext}
\usepackage{amssymb}
\usepackage{nicefrac}       % compact symbols for 1/2, etc.
\usepackage{microtype}      % microtypography
\usepackage{todonotes}
\usepackage{graphicx}
\usepackage{multicol}
\usepackage[ruled,vlined]{algorithm2e}
\usepackage{float}
\usepackage{wrapfig}

\usepackage{mathtools}
\SetKwFunction{Fcreate}{train\_ensemble}
\SetKwFunction{Fsample}{sample\_ensemble}
\SetKwFunction{Fpredict}{ensemble\_predict}
\SetKwProg{Fn}{def}{:}{}
\DontPrintSemicolon

\title{Selective Ensembles for Consistent Predictions}
\iclrfinalcopy
\author{%
  Emily Black \\
  Computer Science Department\\
  Carnegie Mellon University\\
  Pittsburgh, PA 15213 \\
  \texttt{emilybla@cs.cmu.edu} \\
  \And
 Klas Leino  \\
  Computer Science Department\\
  Carnegie Mellon University\\
  Pittsburgh, PA 15213 \\
  \texttt{kleino@cs.cmu.edu} \\
\And
  Matt Fredrikson \\
  Computer Science Department\\
  Carnegie Mellon University\\
  Pittsburgh, PA 15213 \\
  \texttt{mfredrik@cs.cmu.edu} \\
  }

\usepackage{amsmath, amssymb, amsthm}
\usepackage{bm}
\usepackage{xcolor, graphicx}
\usepackage{xspace}
\usepackage{subcaption}
\usepackage{url}
\usepackage{enumerate}
\usepackage{sidecap}
\usepackage{standalone}
\usepackage{caption}
\usepackage{multirow}
\usepackage{dsfont}

\newtheorem{theorem}{Theorem}[section]

\newtheorem{corollary}[theorem]{Corollary}

\newcommand{\pflip}{\ensuremath{p_{\text{flip}}}\xspace}
\newcommand{\pr}{\ensuremath{\mathop{\mathbf{Pr}}}\xspace}

\newcommand{\E}{\mathop{\mathbb{E}}}

\newcommand{\X}{\ensuremath{\mathbf{X}}\xspace}

\newcommand{\Y}{\ensuremath{\mathbf{Y}}\xspace}

\newcommand{\pipeline}{\ensuremath{\mathcal P}\xspace}
\newcommand{\randomness}{\ensuremath{\mathcal S}\xspace}

\DeclareMathOperator*{\argmax}{argmax}

\makeatletter
\newcommand{\oset}[3][0ex]{%
  \mathrel{\mathop{#3}\limits^{
    \vbox to#1{\kern-2\ex@
    \hbox{$\scriptstyle#2$}\vss}}}}
\makeatother

\newcommand{\neqorabs}{\ensuremath{\oset[-0.2em]{_\mathtt{ABS}}{\neq}}}

\newcommand{\modepredictor}{\ensuremath{g_{\pipeline,\randomness}}}
\newcommand{\ensemble}{\ensuremath{\hat g_n(\pipeline, S)}}
\newcommand{\ensembleat}[1]{\ensuremath{\hat g_n(\pipeline, S~;~\alpha,#1)}}

\begin{document}
\maketitle
\begin{abstract}
%!TEX root=paper.tex
Recent work has shown that models trained to the same objective, and which achieve similar measures of accuracy on consistent test data, may nonetheless behave very differently on individual predictions. This inconsistency is undesirable in high-stakes contexts, such as medical diagnosis and finance. We show that this inconsistent%duplicitous 
behavior extends beyond predictions to feature attributions, which may likewise have negative implications for the intelligibility of a model, and one's ability to find recourse for subjects. We then introduce \emph{selective ensembles} to mitigate such inconsistencies by applying hypothesis testing to the predictions of a set of models trained using randomly-selected starting conditions; importantly, selective ensembles can abstain in cases where a consistent outcome cannot be achieved up to a specified confidence level. We prove that that prediction disagreement between selective ensembles is bounded, and empirically demonstrate that selective ensembles achieve consistent predictions and feature attributions while maintaining low abstention rates. On several benchmark datasets, selective ensembles reach zero inconsistently predicted points, with abstention rates as low 1.5\%.
\end{abstract}

\section{Introduction}
%!TEX root=paper.tex

Recent work has drawn attention to the fact that models that appear similar from aggregate quality measures, such as accuracy, often have markedly different behavior at the level of individual predictions~\citep{blackleave2021,marx2019}. 
Further, in deep models, this inconsistency can arise even between closely-related models, such as those arising from different initializations, or from leave-one-out differences in the training data~\citep{blackleave2021,d2020underspecification,mehrer2020individual}. 
This behavior is undesirable in many high-stakes contexts, such as medical applications and credit-approving scenarios, as it may cast doubt on the justifiability of the model's outcome and pose difficulties for reproducibility and comparison. 
% Further, it is especially problematic in situations where models may be routinely re-trained throughout deployment, as these re-trainings may change the model's output, confusing the users or decreasing trust in the model's predictions.  

%he repercussions of this behavior range from confusing the model user, to unjustly or even illegally~\citep{} rejecting a loan-reapplication that has met recourse criteria/
We begin by demonstrating that not only are the predictions of related deep models often dissimilar, but their \textit{feature attributions}~\citep{simonyan2014deep,sundararajan2017axiomatic,leino18influence} are as well (Section~\ref{instability}). 
In particular, we show that there is little connection between a model's gradients, which are the basis for many deep attribution methods, and the labels that it predicts\textemdash models with identical predictions can have arbitrarily different gradients almost everywhere (Proposition~\ref{prop_grad_inst}).
In practice, we show that this result occurs often on common datasets across closely-related models, leading to significant variation in attributions.
This may be undesirable, as feature attributions are commonly used to provide explanations~\citep{simonyan2014deep,sundararajan2017axiomatic,leino18influence}, debug model behavior~\citep{NEURIPS2020_075b051e}, and diagnose problems related to privacy and fairness~\citep{leino2019memorizationprivacy,datta2016algorithmic}.
Beyond these pragmatic concerns, this suggests that the salient factors behind these models' predictions on many points may have little in common, even when models appear to do comparably well on test data.

To address inconsistency in both prediction and attribution, we then turn to ensembling, a well-known approach for reducing predictive variance~\citep{meir1995bias,naftaly1997optimal,lincoln90,fumera2005dynamics,hansen1990neural,krogh1995validation}.
We introduce \emph{selective ensembles}, which leverage a recent result on multinomial rank verification~\citep{hung2019rank}---which has also been used recently for making certifiably-robust predictions~\citep{cohen19certified}---to efficiently mitigate the problem of inconsistency with a probabilistic guarantee.
Given a point to classify, a selective ensemble returns the mode of the class labels predicted on that point, where the mode is sampled over models that vary according to a specified source of randomness in the training process. 
Importantly, if the mode cannot be inferred with sufficient confidence, then the selective ensemble \emph{abstains} from prediction. 
This allows us to bound the probability that these ensembles do not return the true mode prediction (Theorem~\ref{thm:matches_mode}), and by extension, the rate of disagreement between selective ensembles (Corollary~\ref{prop:disagreement}).
In addition, we show that this also bounds the variance component in the ensembles' bias-variance error decomposition~\citep{domingos2000unified} (Corollary~\ref{corr:loss_variance}), providing guidance on how to effectively use of them in practice.

Our experiments show that on seven benchmark datasets, selective ensembles of just ten models either \emph{agree on the entire test data} across random differences in how their constituent models are trained, or abstain at reasonably low rates (1-5\% in most cases; Section~\ref{sec:pred_instability}).
Additionally, we show that simple ensembling doubles the agreement of attributions on key metrics on average, and when the variance of the constituent models is high that selective ensembling further enhances this effect (Section~\ref{sec:grad_instability}).

In summary, our contributions are: \emph{(1)} we show that beyond predictions, feature attributions are not consistent across seemingly inconsequential random choices during learning (Section ~\ref{instability}); \emph{(2)} we introduce \emph{selective ensembling}, a learning method that \emph{guarantees} bounded inconsistency in predictions, (Section ~\ref{ensembles_theory}); and \emph{(3)} we demonstrate the effectiveness of this approach on seven datasets, showing that selective ensembles consistently predict \emph{all} points across models trained with different random seeds or leave-one-out differences in their training data, while also achieving low abstention rates and higher feature attribution consistency.

\section{Notation and Preliminaries}
\label{sec:notation}

We assume a supervised classification setting, with data points $(x, y) \in \X \times \Y$, drawn from data distribution, $\mathcal D$, where $x$ represents a vector of features and $y$ a response. 
In order to capture the effects of arbitrary random events on a learned model---ranging from randomness during training to randomness in the data selection process---we generalize the standard concept of a \emph{learning rule} to that of a \emph{learning pipeline}.
Specifically, a learning pipeline, $\pipeline$, is a procedure that outputs a model, $h : \X \to \Y$, taking as input a random state, $S\sim\randomness$, containing all the information necessary for $\pipeline$ to produce a model (including the architecture, training set, random coin flips used by the learning rule, etc.).
Intuitively, $\randomness$ represents a distribution over random events that might impact the learned model.
For example, $\randomness$ might capture randomness in sampling of the training set, or nondeterminism in the optimization process, e.g., the initialization of parameters, the order in which batches are processed, or the effects of dropout.

In our experiments, we model $\randomness$ to capture two specific types of random choices, namely \emph{(1)} the initial parameters of the model, and \emph{(2)} leave-one-out changes to the training data. 
As the initial parameters of the model tend %in practice 
to be determined by a random seed, we will interchangeably refer to this as the selection of random seed.
More generally, both of these types of choices instantiate a broader class of choices that could be considered \emph{arbitrary}, despite the fact that they may %nonetheless 
impact the predictions~\citep{blackleave2021,marx2019,mehrer2020individual} (Section~\ref{sec:pred_instability}) and explanations (Section~\ref{sec:grad_instability}) of the resulting model.

%We define a \emph{distribution of randomness} $R$ over which a model training process occurs. Some examples of distributions of randomness include be the distribution of all possible random initializations under a certain initializer, e.g. the default Keras glorot uniform initializer~\citep{}, or the uniform distribution of training sets with a one-point difference from some reference training set. When a singular model is trained, it is trained over a random draw from some $R$, e.g., a model randomly chooses its initialization from the distribution of the glorot uniform initializer~\citep{}. We let $h_{S,r}$ be model created by learning rule $h$ on training set $S$ with random draw $r$ from source of randomness $R$. 
% An ensemble model for a learning rule $h$ and dataset $S$ a group of $n$ models each formed from iid runs of $h(S)$. We will refer to the $n$ models that make up an ensemble models as \emph{constituent models}. We denote ensemble models as $E_n(h, S)$. 
% , we refer to a group of $n$ models each formed with different random draw from $R$. We refer to this as $E_n(h, S, R)= \{h_{S,r_i}, ...h_{S,r_k} \}$, where $\{r_i..r_k\}$ are random draws over $R$, and $|\{r_i..r_k\}|=n$. 
\section{Instability of Feature Attributions in Deep Models} 
\label{instability}
%!TEX root=paper.tex

Before we consider mitigating predictive inconsistency with ensembling, we first demonstrate that models' inconsistency across random choices in training is exhibited not only through its predictions, but through its \emph{feature attributions} as well. 
Feature attributions refer to numeric scores generated for some set of a model's features---most commonly the model's input features---which are meant to connote how important each feature is in generating the model's prediction.
Feature attributions are a commonly used as a tool for explaining model behavior~\citep{simonyan2014deep,leino18influence,sundararajan2017axiomatic,NEURIPS2020_075b051e} localized to given set of inputs.
Thus, inconsistent feature attributions between models suggest the models differ in the \emph{process} by which they arrive at their predictions, even if the predictions are the same.
% perhaps hints at the fact that there is inconsistency in the \emph{method} by which a model makes a decision on a given point across arbitrary differences to the model.
% Beyond this, differences between attributions of models with inconsequential differences may lead to user confusion, or even distrust of a model's decision process.
% Differences in the decision process that arise as a result of arbitrary choices made throughout a learning pipeline as a model is retrained and redeployed may lead to user confusion, or even distrust of a model's decision process.

\begin{figure*}[!tbp]
\resizebox{\textwidth}{!}{%
  \centering
  \begin{minipage}[b]{0.5\textwidth}
    \includegraphics[width=\textwidth]{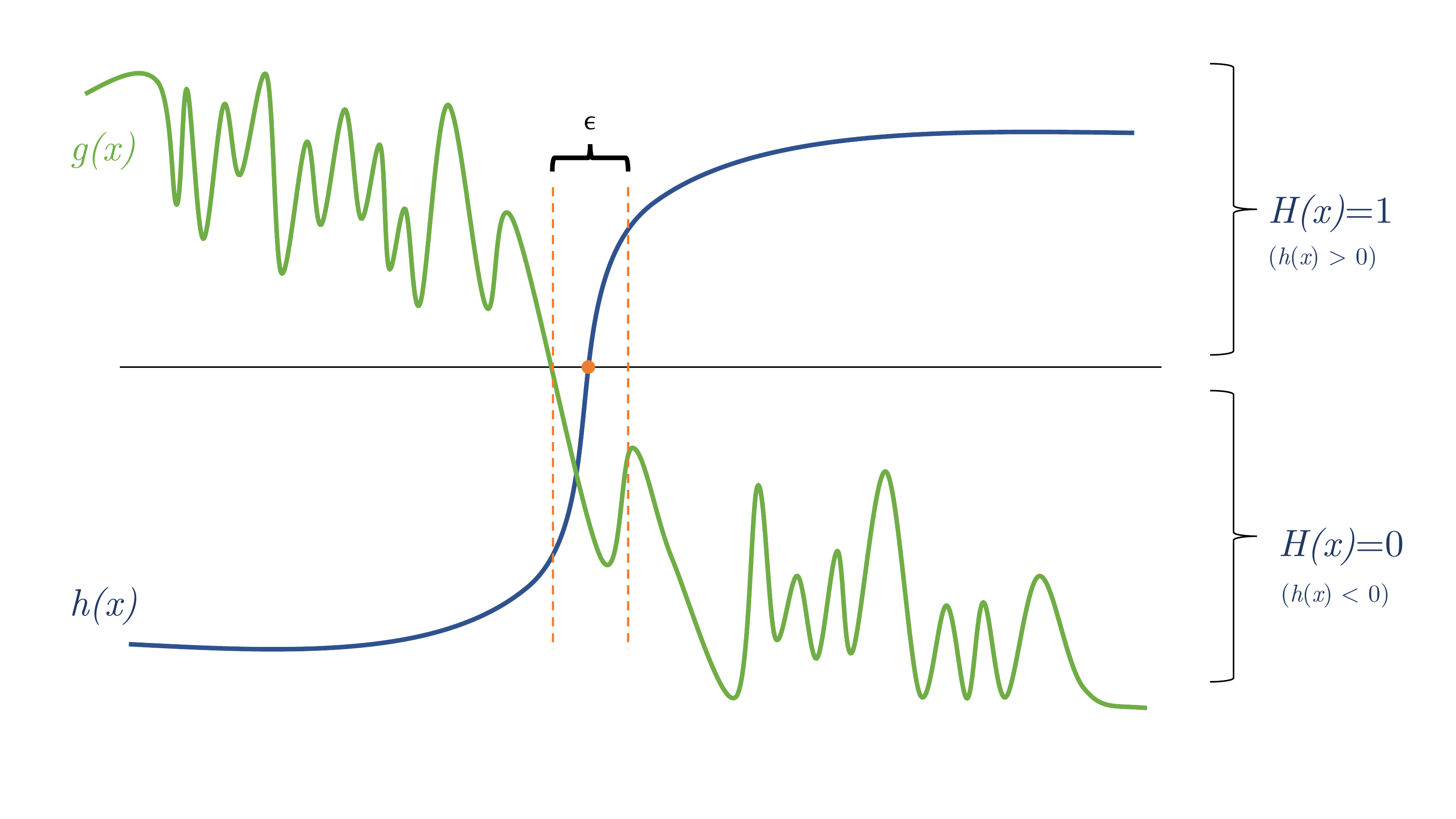}
  \end{minipage}
  \hfill
  \begin{minipage}[b]{0.5\textwidth}
    \includegraphics[width=\textwidth]{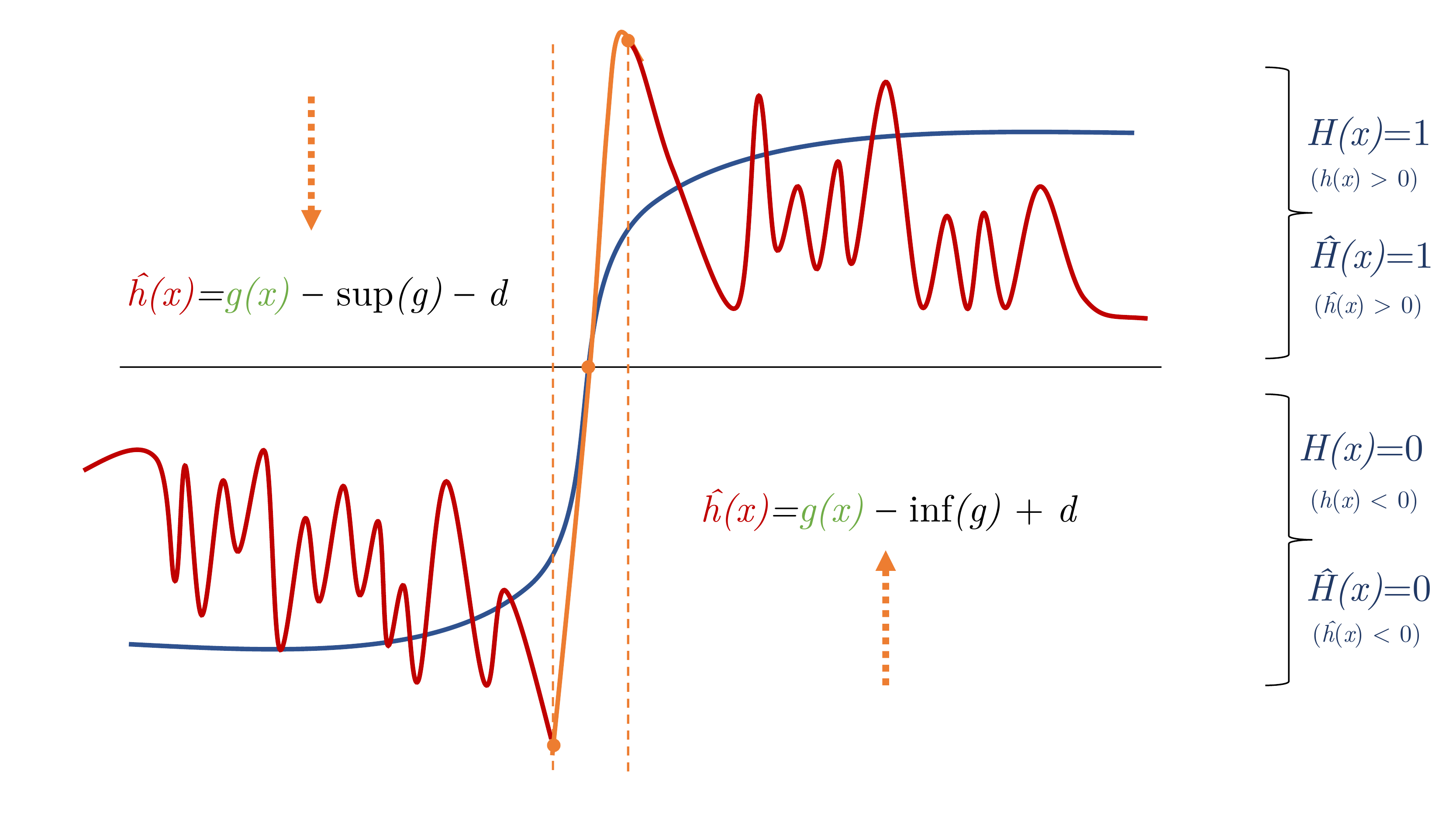}
  \end{minipage}
 }
 \caption{Intuitive illustration of how two models which predict identical classification labels can have arbitrary gradients. To show this, given a binary classifier $H$ and an arbitrary function $g$, we construct a classifier $H'$ that predicts the same labels as $H$, yet has gradients equal to $g$ almost everywhere. We formally state this result in Theorem~\ref{prop_grad_inst}.}
 \label{expl_proof_int}
\end{figure*}

In deep models, many of the most popular attribution methods are based on the model's gradients at or around a given point~\citep{simonyan2014deep,sundararajan2017axiomatic}. 
Accordingly, we will focus on the stability of gradients, and show via analysis and experiment that they are not stable in conventional deep models.
%Before we consider the mitigation method of selective ensembling, we first demonstrate that the problem of inconsistency of deep models across arbitrary changes such as random seed and one-point differences in the training set extends beyond the models predictions, and in to its gradient-based explanations. The gradient of a point in a model with respect to the input space is a proxy for understanding which features were most important in making the model's decision. Thus, this inconsistency of saliency maps perhaps hints at the fact that there is inconsistency in \emph{the method by which a model makes a decision on a given point} across arbitrary differences to the model.
First, we motivate our results by showing that even two deep models that predict the same labels on all points may %nonetheless 
have arbitrarily different gradients almost everywhere. 
Later, in our empirical evaluation (Section~\ref{sec:grad_instability}), we demonstrate the extent of the differences between Saliency Maps~\citep{simonyan2014deep} (i.e., input gradients) of deep networks %trained on standard benchmark datasets 
even when the randomness of the learning pipeline is controlled to allow only one-point differences in the training set or differences in the random seed.
%Then, we demonstrate the extent of the differences between saliency maps (i.e. gradients) of deep learning models which differ only on random seed or a one-point difference in the training set.

\paragraph{Predictions with Arbitrary Gradients.}
We show that even deep models that predict the exact same labels on all points cannot necessarily be expected to have the same, or even similar, gradients;
in fact, given a binary classification model $h$, we can construct a model $\hat h$ which predicts the same labels as $h$, but has arbitrarily different gradients everywhere except an arbitrarily small region around the boundary of $h$ (Theorem~\ref{prop_grad_inst}).

\begin{theorem}
\label{prop_grad_inst}
Let $H : \X \rightarrow \{-1,1\} = \mathrm{sign}(h)$ be a binary classifier and
$g: \mathbb{R}^n \rightarrow \mathbb{R}$ be an unrelated function that is bounded from above and below, continuous, and piecewise differentiable.
Then there exists another binary classifier $\hat H = \mathrm{sign}(\hat h)$ such that for any $\epsilon > 0$,
$$
\forall x\in\X~.\hspace{2em}
\text{1.}~~\hat H(x) = H(x)\hspace{2em}
\text{2.}~~\inf_{x' : H(x')\neq H(x)}\Big\{||x - x'||\Big\} > \nicefrac{\epsilon}{2}~~\Longrightarrow~~\nabla \hat h(x) = \nabla g(x)$$
\end{theorem}

The proof of Theorem~\ref{prop_grad_inst} is given in Appendix~\ref{app:saliency_proof} in the supplementary material.
The proof is by construction of $\hat h$; a sketch giving the intuition behind the construction is provided in Figure~\ref{expl_proof_int}.
In short, we first partition the domain into contiguous regions that are given the same label by $H$.
We then construct $\hat h$ from $g$ by adjusting $g$ to lie above or below the origin to match the prediction behavior of $h$ in each region.
As these transformations merely shift $g$ by a constant in each region, they do not change $\nabla g$ except near decision boundaries, where it is necessary to move across the origin.

\paragraph{Observations.}
The intuition stemming from Theorem~\ref{prop_grad_inst} is that a model's gradients at each point are largely disconnected from the labels it predicts on a distribution. 
As models that make identical predictions are likely to have similar loss on a given dataset, this theorem points to the possibility that models of similar objective quality may still have arbitrarily different gradients. 
In Section ~\ref{sec:grad_instability}, we demonstrate that this outcome is not only \emph{possible}, but that it occurs in real models%In particular, models that achieve very similar levels of accuracy often have inconsistent feature attributions
---for example, on the German Credit dataset predicting credit risk, on average, individual models with similar accuracy agree on \emph{less than two out of the five most important features} influencing their decision.

\section{Selective Ensembling}
\label{ensembles_theory}
%!TEX root=paper.tex

% As noted in the section above, deep models can be relatively inconsistent in their decisions and decision procedures across arbitrary, insignificant changes. However, there may be some contexts where we need a model to be consistent across small changes such as random seed, but we still want the capabilities of a deep model---for example, in some deployment context for an algorithm used in a medical context, which is retrained over time to ensure continued accuracy, or privacy. How might we accomplish this?
% First, we consider how a model could make perfectly stable predictions over a source of randomness---and define such a hypothetical model, the \emph{majority prediction function} over a source of randomness. Then, we describe how to approximate this function in practice with \emph{selective ensemble models}. Finally, we demonstrate their efficacy.
% \subsection{Ensemble Models and Majority Predictions}
The results of Section~\ref{instability} suggests that models that are retrained and redeployed, %even under similar circumstances, may nonetheless 
may exhibit substantially different behavior from their previous iterations.
We build on the approach of ensembling for variance reduction by showing how these differences in behavior can be bounded via \emph{selective ensembling}.
However, whereas prior work which finds that \emph{more diversity} among the constituent networks is beneficial for reducing overall error~\citep{krogh1995validation,hansen1990neural,maclin1995combining,Opitz96}, our goal is to minimize, or at least place strict bounds on, the variance component.
We show that ideas from robust classification, and in particular \emph{randomized smoothing}~\citep{cohen19certified}, which stem from recent results on multinomial hypothesis testing~\citep{hung2019rank}, can be used to enforce such a bound.

\paragraph{Mode Predictor.} 
We may view the image of the learning pipeline, $\pipeline$, as a distribution over possible models induced by applying $\pipeline$ to the random state, $S\sim\mathcal S$.
The \emph{mode prediction} on an input $x$, with respect to $\randomness$, is the expected label that would be predicted on $x$ by models drawn from this distribution.
More formally, we define the \emph{mode predictor}, $g_{\pipeline,\randomness}$ for a pipeline, $\pipeline$, and random state distribution, $\randomness$, as given by Equation~\ref{eq:mode_predictor}.
\begin{equation}
\label{eq:mode_predictor}
g_{\pipeline,\randomness}(x) = \argmax_{y\in\Y}\left\{\E_{S\sim\randomness}\Big[\mathds{1}[\pipeline(S~;~x) = y]\Big]\right\}
\end{equation}

Note that while $g_{\pipeline,\randomness}$ is deterministic, and is therefore not sensitive to a specific state drawn from $\randomness$, it does not necessarily produce the ground truth label on all inputs---some learning pipelines may converge to a stable loss minimum that misclassifies certain points.

\paragraph{Approximation via Ensembling.} 
An explicit representation of the true mode predictor is, of course, unattainable---the non-convex loss surface of deep models and the complex interactions between the learning pipeline and the distribution of random states makes the expectation in Equation~\ref{eq:mode_predictor} infeasible to compute analytically. 
However, we can approximate $g_{\pipeline,\randomness}(x)$ by computing the empirical mode prediction on $x$ over a random sample of models produced by i.i.d. draws from $\pipeline(S)$. 
% In other words, we can approximate the mode predictor by predicting the class that wins a plurality vote across an ensemble of models.
% ... but we would like to quantify the extent to which our empirical measurement matches the true mode...
% Moreover, for a fixed-size ensemble, there may still be points on which we cannot be certain...
% We deal with the former problem 
But although ensembles with sufficiently many constituent models will more reliably output the mode prediction, for any fixed-size ensemble there will remain points on which the margin of the plurality vote is small enough to ``flip'' to runner-up in some set of nearby ensembles that differ on a subset of their constituents; in other words, these ensembles will not predict the mode prediction.

To rigorously bound the rate at which the ensemble will differ from the mode prediction, we allow the ensemble to \emph{abstain} on points where the constituent predictions indicate a statistical toss-up between the two most likely classes. 
We call ensembles that may abstain in this way \emph{selective ensembles}, borrowing the terminology from selective classification~\citep{el2010foundations}.
We can think of of abstention as a means of flagging unstable points on which the selective ensemble cannot accurately determine the mode prediction; whether this should be interpreted as a failed attempt at classification is an application-specific consideration. 
% We will show that the fraction of points that are unstable but \emph{not} flagged by the selective ensemble can be bounded (Theorem~\ref{thm:matches_mode}).

\begin{figure}
\centering
\small
\begin{minipage}{0.47\textwidth}
\begin{algorithm}[H]
\vspace{0.5em}
\Fn{\Fcreate{$\pipeline,~~S\sim\randomness^n,~~n$~}}{
	\textbf{return} $\{\pipeline(S_i)~~\text{for}~~i\in[n]\}$\;
}
\vspace{2em}
\Fn{\Fsample{$\pipeline,~~\randomness,~~n$~}}{
	$S~~\gets~~\mathtt{sample\_iid}(\randomness^n)$\;
	\textbf{return} $\Fcreate(\pipeline,~S,~n)$\;
}
\vspace{1em}
\caption{Selective Ensemble Creation}
\label{alg:create_ensemble}
\end{algorithm}
\end{minipage}\quad %\hspace{0.05\textwidth}%
\begin{minipage}{0.48\textwidth}
\begin{algorithm}[H]
\Fn{\Fpredict{$\hat g_n(\pipeline,S),~~\alpha,~~x$~}}{
	$Y~~\gets~~\sum_{h \in \hat g_n(\pipeline,S)} \mathtt{one\_hot}(h(x))$\;
	$n_A,~n_B~~\gets~~\mathtt{top\_2}(Y)$\;
	\eIf{$\mathtt{binom\_p\_value}(n_A,~n_A+n_B,~0.5)\leq \alpha$}{
		\textbf{return} $\mathtt{argmax}(Y)$\;
	}{
		\textbf{return} $\mathtt{ABSTAIN}$\;
	}
}
\caption{Selective Ensemble Prediction}
\label{prediction_algo}
\end{algorithm}
\end{minipage}
\end{figure}

Selective ensembles of $n$ models predict according to the following procedure.
First, the predictions of each of the $n$ models in the ensemble are collected. 
The constituent models are derived from $n$ i.i.d. samples of $\pipeline(S)$ from $\randomness$, as described in Algorithm~\ref{alg:create_ensemble}.
From these predictions, we perform a two-sided statistical test to determine if the mode prediction was selected by a statistically significant majority of the constituent models.
%In other words, we attempt to refute the hypothesis that the number of constituent models selecting the mode prediction was drawn from a binomial distribution with $p=0.5$.
If the statistical test succeeds, we return the empirical mode prediction; otherwise we abstain from predicting. 
Pseudocode for this prediction procedure is given in Algorithm~\ref{prediction_algo}.
We will denote by $\hat{g}_n(\pipeline,S)$ (for $S\sim\randomness^n$) the output of $\Fcreate$ in Algorithm~\ref{alg:create_ensemble}, and by $\hat g_n(\pipeline, S~;~\alpha, x)$ prediction produced by $\Fpredict$ in Algorithm~\ref{prediction_algo} on $\hat{g}_n(\pipeline,S)$.

Because of their ability to abstain from prediction, we can prove that with probability at least $1 - \alpha$, a selective ensemble will either return the true mode prediction or abstain, where $\alpha$ is a chosen threshold for the statistical test to prevent prediction in the case of a toss-up. 
In other words, on any point on which it does not abstain, a selective ensemble will disagree with the mode predictor, $g_{\pipeline,\randomness}$, with probability at most $\alpha$, as stated formally in Theorem~\ref{thm:matches_mode}.

The statement of Theorem~\ref{thm:matches_mode} make use of the relation, $\neqorabs$, where $y_1 \neqorabs y_2$ if and only if $y_1\neq\mathtt{ABSTAIN}$ and $y_2\neq\mathtt{ABSTAIN}$ and $y_1\neq y_2$.
That is, $\neqorabs$ captures disagreement between non-rejected predictions.

\begin{theorem}
\label{thm:matches_mode}
Let $\pipeline$ be a learning pipeline, and let $\randomness$ be a distribution over random states.
Further, let $g_{\pipeline,\randomness}$ be the mode predictor, let $\hat{g}_n(\pipeline,S)$ for $S\sim\randomness^n$ be a selective ensemble, and let $\alpha \geq 0$.
Then,
$$
\forall x\in\X~~.~~
\pr_{S\sim\randomness^n}\Big[
\hat{g}_n(\pipeline,S~;~\alpha, x) \neqorabs g_{\pipeline,\randomness}(x)
\Big] \leq \alpha
$$ 
\end{theorem}
The proof (Appendix~\ref{appendix:proofs}) relies on a result from Hung and Fithian~\citep{hung2019rank} which bounds the probability that a set of votes does not return the true plurality outcome, and we apply it in a similar fashion to how it is used for making robust predictions in Randomized Smoothing~\citep{cohen19certified}. 
% \begin{proof}
% $En(h, S, R, n)$ is an ensemble of $n$ models. By the definition of the Predict algorithm,  $En(h, S, R, n)$ gathers a vector of class counts the prediction for $x$ from each model in the ensemble. Let the class with the highest count be $c_A$, with counts $n_A$, and the class with the second highest count be called $c_B$, with counts $n_B$. In order for the ensemble to predict, the ensemble models runs a two-sided hypothesis test to ensure that $Pr[n_A \sim \text{Binomial}(n_A+n_B, 0.5)] < \alpha$, i.e. that $A$ is the true majority prediction over $R$. See that
%     	 $$ P[g_{h,S,R}(x)\neq c_A \land En(h, S, R, n)(x)=c_A ] $$
%     	$$ = P[g_{h,S,R}(x) \neq c_A] P[En(h, S, R, n) \text{does not abstain} | g_{h,S,R}(x) \neq c_A]$$
%     	$$\leq P[En(h, S, R, n)\text{ does not abstain} |g_{h,S,R}(x) \neq c_A]$$

%     	By Hung and Fithian\citepp{}, we have that 
%     	 $$\leq P[En(h, S, R, n) \text{ does not abstain} | g_{h,S,R}(x) \neq c_A] = \alpha$$
%     	 Thus,
%     	$$P[g_{h,S,R}(x) \neq c_A \land En(h, S, R, n)(x)=c_A ] \leq \alpha $$
% \end{proof}

Theorem~\ref{thm:matches_mode} states that the probability that a selective ensemble makes a prediction that does not match the mode prediction is small.
However, one possible means of ensuring this is by not providing a prediction in the first place, i.e., if the selective ensemble abstains.
Thus, the \emph{abstention rate} is necessary to quantify the fraction of points on which the mode prediction will actually be produced.

In the 0-1 loss bias-variance decomposition of \citet{domingos2000unified}, the variance component of a classifier's loss is defined as the expected loss relative to the mode prediction (in our case, taken over the randomness in $\randomness$).
Thus, Theorem~\ref{thm:matches_mode} leads to a direct bound on this component, assuming a bound, $\beta$, on the abstention rate. 
This is formalized in Corollary~\ref{corr:loss_variance}.

\begin{corollary}
\label{corr:loss_variance}
Let $\pipeline$ be a learning pipeline, and let $\randomness$ be a distribution over random states.
Further, let $g_{\pipeline,\randomness}$ be the mode predictor, let $\hat{g}_n(\pipeline,S)$ for $S\sim\randomness^n$ be a selective ensemble.
Finally, let $\alpha \geq 0$, and let $\beta \geq 0$ be an upper bound on the expected abstention rate of $\hat{g}_n(\pipeline,S)$.
Then, the expected \emph{loss variance}, $V(x)$, over inputs, $x$, is bounded by $\alpha + \beta$.
That is,
$$
\E_{x\sim\mathcal D}\Big[V(x)\Big] 
= \E_{x\sim\mathcal D}\Bigg[~
	\Pr_{S\sim\randomness^n}\Big[ \hat{g}_n(\pipeline,S~;~x) \neq g_{\pipeline,\randomness}(x) \Big]
~\Bigg] 
\leq \alpha + \beta
$$ 
\end{corollary}

\begin{figure}
\centering
\resizebox{\textwidth}{!}{%
\includegraphics{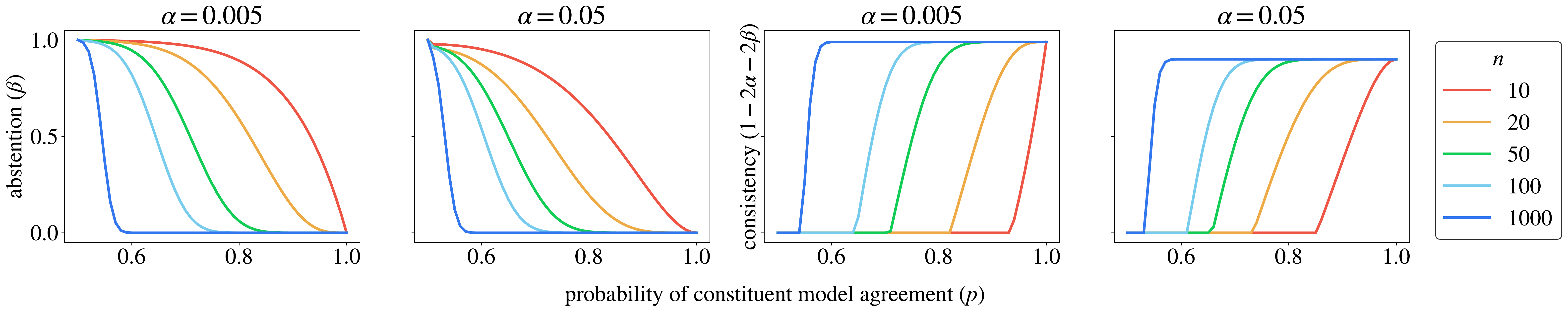}}
\caption{The left two plots show abstention rates as a function of the underlying probability of agreement among models over \randomness, i.e., the probability that any given model will return the mode prediction, with plots denoting varying numbers of constituent models. The right two graphs demonstrate the relationship between consistency of the ensemble models as given by Corollary~\ref{prop:disagreement}.}
\label{fig:abstention}
\end{figure}

\paragraph{Consistency of Selective Ensembles.} 
Using the result from Theorem~\ref{thm:matches_mode}, we can also address the original problem raised: that deep models often disagree on their predictions due to arbitrary random events over the training pipeline.
We show that, given a bound, $\beta$, on the abstention rate, the probability that two selective ensembles disagree in their predictions is bounded by $2(\alpha + \beta)$ (Corollary~\ref{prop:disagreement}). 
Intuitively, this suggests that the predictions of selective ensembles are more stable over different instantiations of the random decisions captured by $\randomness$ compared to individual models.
\begin{corollary}
\label{prop:disagreement}
Let $\pipeline$ be a learning pipeline, and let $\randomness$ be a distribution over random states.
Further, let $\hat{g}_n(\pipeline,S)$ for $S\sim\randomness^n$ be a selective ensemble.
Finally, let $\alpha \geq 0$, and let $\beta \geq 0$ be an upper bound on the expected abstention rate of $\hat{g}_n(\pipeline,S)$.
Then,
$$
\E_{x\sim\mathcal D}\Bigg[~
\pr_{S^1,S^2\sim\randomness^n}\Big[
\hat{g}_n(\pipeline,S^1~;~\alpha, x) \neq \hat{g}_n(\pipeline,S^2~;~\alpha, x)
\Big] 
~\Bigg] 
\leq 2(\alpha + \beta)
$$
\end{corollary}
Corollary~\ref{prop:disagreement} tells us that the agreement between any two selective ensembles is at least $1 - 2(\alpha + \beta)$.
For a fixed $n$, decreasing $\alpha$ will lead to a higher abstention rate.
Thus in order for $\alpha$ \emph{and} $\beta$ to both be small, as would be necessary for a high fraction of consistently-predicted points, we may require a large number of constituent models, $n$.
Figure~\ref{fig:abstention} illustrates the trade-off between $\alpha$, $\beta$, and $n$, depending on the base level of agreement of the constituent models.
In Section~\ref{eval}, we show empirically that even with small values of $n$, abstention rates of selective ensembles are reasonably low in practice.

In summary, selective ensembles accomplish three primary things: (1) they identify points on which the mode prediction cannot be determined, (2) they bound the fraction of points that can be inconsistently predicted, and (3) they provide a means of reliably inferring the mode prediction when the abstention rate can be kept sufficiently low.

\section{Evaluation}
\label{eval}
%!TEX root=paper.tex
\begin{table}[t]
\resizebox{\textwidth}{!}{%
  \begin{tabular}{l|lllllll}
% \toprule
         & \multicolumn{6}{c}{\emph{mean accuracy $\pm$ standard deviation}} \\
Randomness&  Ger. Credit & Adult  &  Seizure & Warfarin  & Tai. Credit & FMNIST &  Colon\\
\midrule
RS & $.730 \pm .020$ & $.842 \pm 1e-3$  & $.973 \pm 2e-3$ & $.686 \pm 3e-3$ & $.820 \pm 1e-3$  & $.916 \pm 3e-3$ & $.927 \pm 2e-3$\\
LOO  & $.729 \pm .012$  & $.843 \pm 7e-4$ & $.976 \pm 2e-3$ & $.686 \pm 2e-3$ & $.820 \pm 1e-3$ & $.917 \pm 8e-4$ & $.926 \pm 3e-3$\\
\bottomrule
\end{tabular}
}
\caption{Mean accuracy over $500$ models trained over changes to random initialization and leave-one-out differences in training data. 
%While the variance is nearly always very small, 
German Credit stands as an outlier due to its small sample size ($|D|=800$). }
\label{accs}
\end{table}

\begin{table}[t]
\resizebox{\textwidth}{!}{%
  \begin{tabular}{ll|lllllll}
% \multicolumn{4}{c}{portion of Taiwanese Credit test data with $\pflip > 0$} \\
%Randomness& $n$ & Mean & Std \\
& & \multicolumn{5}{c}{\emph{mean of portion of test data with $\pflip > 0$}} \\
Randomness& $n$ & Ger. Credit & Adult  &  Seizure & Tai. Credit & Warfarin  & FMNIST &  Colon\\
\midrule
 %& & Mean & Std.\\
RS & 1 & $.570 $ & $.087  $ & $.060$  &   $.082 $ & $.098 $ & $.061 $  & $.037 $   \\
RS & (5, 10, 15, 20) & ~$0.0$  & ~$0.0$ & ~$0.0$  & ~$0.0$  & ~$0.0$  & ~$0.0$  & ~$0.0$   \\
LOO & 1 &  $.262 $ & $.063 $ & $.031 $ & $.031 $ & $.033 $ & $.034 $ & $.042 $ \\
LOO &(5, 10, 15, 20) & ~$0.0$  & ~$0.0$ & ~$0.0$  & ~$0.0$  & ~$0.0$  & ~$0.0$  & ~$0.0$  \\
\bottomrule
\end{tabular}
}
\caption{Percentage of points with disagreement between at least one pair of models ($\pflip > 0$) trained with different random seeds (RS) or leave-one-out differences (LF) in training data, for singleton models ($n=1$) and selective ensembles ($n > 1$). Results are averaged over 10 runs of creating 24 selective ensemble models, standard deviations are in Appendix~\ref{appendix:more_ens_results}. Selective ensemble results are together, as there is no disagreement.}
\label{selective_ens_flipping}
\end{table}

\begin{table}[t]
\resizebox{\textwidth}{!}{%
  %!TEX root=../paper.tex

\begin{tabular}{ll|r@{\hskip 0pt}l@{\hskip 0pt}lr@{\hskip 0pt}l@{\hskip 0pt}lr@{\hskip 0pt}l@{\hskip 0pt}lr@{\hskip 0pt}l@{\hskip 0pt}lr@{\hskip 0pt}l@{\hskip 0pt}lr@{\hskip 0pt}l@{\hskip 0pt}lr@{\hskip 0pt}l@{\hskip 0pt}l}
% \toprule
         & \multicolumn{21}{c}{\emph{accuracy (abstain as error) / abstention rate / non-selective accuracy}} \\
\randomness & $n$ & \multicolumn{3}{c}{Ger. Credit} & \multicolumn{3}{c}{Adult}  &  \multicolumn{3}{c}{Seizure} & \multicolumn{3}{c}{Warafin}  & \multicolumn{3}{c}{Tai. Credit} & \multicolumn{3}{c}{FMNIST} & \multicolumn{3}{c}{Colon} \\
\midrule
RS & 5 & $0.0$ & $/ 1.0$ & $/ .745$ & $0.0$ & $/ 1.0$ & $/ .842$ & $0.0$ & $/ 1.0$ & $/ .975$ & $0.0$ & $/ 1.0$ & $/ .688$ & $0.0$ & $/ 1.0$ & $/ .822$ & $0.0$ & $/ 1.0$ & $/ .919$ & $0.0$ & $/ 1.0$ & $/ .927$ \\
RS & 10 & $.576$ & $/ .291$ & $/ .746$ & $.820$ & $/ .043$ &$/ .843$ & $.960$ & $/ .026$ & $/ .975$ & $.660$ & $/ .050$ & $/ .688$ & $.800$ & $/ .039$ & $/ .822$ & $.888$ & $/ .059$ & $/ .920$ & $.914$ & $/ .032$ & $/ .928$\\
RS & 15 & $.636$ & $/ .205$ & $/ .750$ & $.827$ & $/ .032$  & $/ .842$ & $.965$ & $/ .018$  &$/.975$ &$.668$ & $/ .037$& $/ .688$ & $.807$ & $/ .028$  & $/ .822$ & $.897$ & $/ .042$ & $/ .920$ & $.919$ & $/ .023$ & $/ .928$\\
RS & 20 & $.664$ & $/ .165$ & $/ .747$ & $.830$ & $/ .024$ &$/ .842 $ & $.967$ & $/ .014$ &  $/ .975$ & $.670$ & $/ .031$& $/ .688$ & $.810$ & $/ .023$  &  $/ .822$ & $.902$ & $/ .036$ & $/ .920$ & $.921$ & $/ .019$ & $/ .938$\\
\midrule
LOO & 5 & $0.0$ & $/ 1.0$ & $/ .728$ & $0.0$ & $/ 1.0$ & $/ .844$ & $0.0$ & $/ 1.0$ & $/ .978$ & $0.0$ & $/ 1.0$ & $/ .685$ & $0.0$ & $/ 1.0$ & $/ .821$ & $0.0$ & $/ 1.0$ & $/ .918$ & $0.0$ & $/ 1.0$ & $/ .927$ \\
LOO & 10  & $.653$ & $/ .151$ & $/.728$ & $.827$ & $/ .032$ & $/ .844$& $.962$ & $/ .027$ & $/ .978$& $.677$ & $/ .018$& $/ .685$ & $.812$ & $/ .017$ & $/ .821$ & $.909$ & $/ .020$  & $/ .918$ & $.912$ & $/ .036$ & $/ .927$\\
LOO & 15  & $.678$ & $/ .105$ & $/.733$ &$.832$ & $/ .012$ & $/ .844$& $.968$ & $/ .019$ & $/ .979$& $.679$ & $/ .013$& $/ .685$ & $.814$ & $/ .013$ & $/ .821$ & $.910$ & $/ .016$ & $/ .917$ & $.916$ & $/ .027$ & $/ .927$\\
LOO & 20  & $.689$ & $/ .079$  & $/.730$ & $.834$ & $/ .018$& $/ .843$& $.970$ & $/ .015$ & $/ .979$ & $.680$ & $/ .011$& $/ .685$ & $.815$ & $/ .010$ & $/ .821$ & $.912$ & $/ .012$ & $/ .918$ & $.919$ & $/ .023$ & $/ .927$\\
\bottomrule
\end{tabular}
}
\caption{Accuracy and abstention rate of selective ensembles, along with the accuracy of non-selective (traditional ensembles) with $n \in \{5, 10, 15, 20\}$ constituents. %Results are presented as follows: selective accuracy/ abstention rate, non-selective accuracy.
Results are averaged over 24 randomly selected models; standard deviations are given in Table~\ref{app:tab:ens_acc_abs} in Appendix~\ref{appendix:more_ens_results}}
\label{ens_accs_abs}
\end{table}

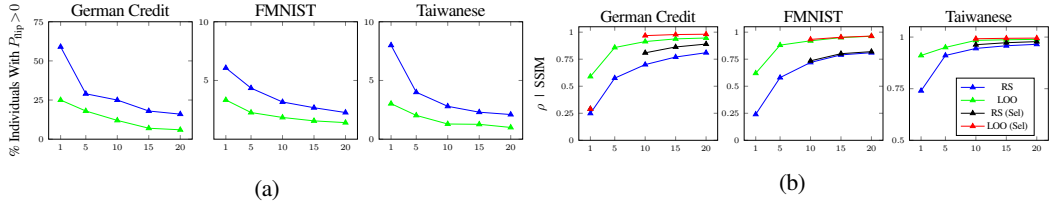
\begin{figure}[t]
\begin{subfigure}{0.5\textwidth}
\resizebox{\textwidth}{!}{%
% !TEX root = ../paper.tex

%\documentclass{standalone}

% \usepackage{pgfplots}
% \pgfplotsset{compat=1.16}
% \usepgfplotslibrary{groupplots}
% \usetikzlibrary{patterns}

% \begin{document}
\begin{tikzpicture}
\begin{groupplot}[%
	group style = {group size=7 by 1, horizontal sep=15pt, vertical sep=20pt}, 
	% classes={
	% 	a={mark=triangle*,blue},
	% 	b={mark=triangle*,green},
	% 	c={mark=triangle*,red},
	% 	d={mark=triangle*, black}
	% },
	footnotesize,
	xtick={1, 5, 10, 15, 20},
	%ytick={0, 5, 10, 20, 25, 30, 35, 40, 45},
	% yticklabel style={
	% 	/pgf/number format/precision=3,
	% 	/pgf/number format/fixed
	% },
	ymin=0,
	legend style={nodes={scale=0.8}},
	height=1.75in,
	width=2.0in,
	label style={font=\normalsize},
	title style={font=\large},
	tick label style={font=\tiny}
]

% gc graph 2
\nextgroupplot[ymax=75, ytick={0.0, 25, 50, 75}, ylabel={\% Individuals With $P_{\text{flip}}>0$}
, title={German Credit}]

\addplot[color=blue,mark=triangle*,blue] coordinates {
(1.0, 59)
(5.0, 29)
(10.0,25)
(15.0, 18)
(20, 16)
};

\addplot[color=green,mark=triangle*,green] coordinates {
(1.0, 25)
(5.0, 18)
(10.0, 12)
(15.0, 07)
(20, 06)
};

\nextgroupplot[ymax=10, ytick={5,10}, xtick={1, 5, 10, 15, 20}, title={FMNIST}]

\addplot[color=blue,mark=triangle*,blue] coordinates {
	(1.0, 6.07)
(5.0, 4.35)
(10.0, 3.168)
(15.0, 2.68)
(20.0, 2.27)
};

\addplot[color=green,mark=triangle*,green] coordinates {
(1.0, 3.34)
(5.0, 2.27)
(10.0, 1.85)
(15.0, 1.56)
(20.0, 1.41)
};

% \addplot[color=yellow,mark=triangle*,yellow] coordinates {
% 	(0.0, 0.0)
% (0.1, 0.0)
% (0.2, 0.0)
% (0.30000000000000004, 0.0)
% (0.4, 0.0)
% (0.5, 0.0)
% (0.6000000000000001, 0.0)
% (0.7000000000000001, 0.0)
% (0.8, 0.0)
% (0.9, 0.0)

% };

% TODOOOOOOOOO
\nextgroupplot[ymax=10, ytick={0.0, 5, 10}, title={Taiwanese}]
%deep base
\addplot[color=blue,mark=triangle*,blue] coordinates {
	(1.0, 8.0)
(5.0, 4.0)
(10.0, 2.8)
(15.0, 2.3)
(20.0, 2.1)
};

\addplot[color=green,mark=triangle*,green] coordinates {
(1.0, 3.0266666666666667)
(5.0, 2.0266666666666665)
(10.0, 1.2933333333333333)
(15.0, 1.2666666666666666)
(20.0, 1.0)
};

%lin seed
% \addplot[color=yellow, mark=triangle*,yellow] coordinates {
% (0.0, 65.9)
% (0.1, 41.9)
% (0.2, 18.5)
% (0.30000000000000004, 4.3)
% (0.4, 0.4)
% };

% % TODOOOOOOOOO
% \nextgroupplot[ymax=10, ytick={0, 5, 10}, title={Warfarin}]
% %deep base
% \addplot[color=blue,mark=triangle*,blue] coordinates {
% 	(1.0, 0)
% (5.0, 0)
% (10.0, 0)
% (15.0, 0)
% (20.0, 0)

% };

% \addplot[color=green,mark=triangle*,green] coordinates {
% (1.0, 2.8)
% (5.0, 1.6)
% (10.0, 1.3)
% (15.0, 1.1)
% (20.0, 1.0)
% };

% \nextgroupplot[ymax=10, ytick={0, 5, 10}, title={Colon}]
% %deep base
% \addplot[color=blue,mark=triangle*,blue] coordinates {
% 	(1.0, 6.0)
% (5.0, 1.7)
% (10.0, 1.2)
% (15.0, 1.1)
% (20.0, 0.9)
% };

% \addplot[color=green,mark=triangle*,green] coordinates {
% 	(1.0, 6.8)
% (5.0, 2.2)
% (10.0, 1.3)
% (15.0, 1.0)
% (20.0, 0.96)
% };

%lin seed
% \addplot[color=yellow, mark=triangle*,yellow] coordinates {
% (0.0, 65.9)
% (0.1, 41.9)
% (0.2, 18.5)
% (0.30000000000000004, 4.3)
% (0.4, 0.4)
% };

\end{groupplot}

%\legend{Deep, Linear, Adv., Smooth};

\end{tikzpicture}
%\end{document}
}
\caption{}
\label{no_abst}
\end{subfigure}%
\begin{subfigure}{0.5\textwidth}
\resizebox{\textwidth}{!}{%
% !TEX root = ../paper.tex

%\documentclass{standalone}

% \usepackage{pgfplots}
% \pgfplotsset{compat=1.056}
% \usepgfplotslibrary{groupplots}
% \usetikzlibrary{patterns}

% \begin{document}
\begin{tikzpicture}
\begin{groupplot}[%
	group style = {group size=4 by 1, horizontal sep=19pt, vertical sep=20pt},%{group size=7 by 1, horizontal sep=15pt, vertical sep=20pt}, 
	% classes={
	% 	a={mark=triangle*,blue},
	% 	b={mark=triangle*,green},
	% 	c={mark=triangle*,red},
	% 	d={mark=triangle*, black}
	% },
	footnotesize,
	xtick={1, 5, 10, 15, 20},
	%ytick={0, 5, 10, 20, 25, 30, 35, 40, 45},
	% yticklabel style={
	% 	/pgf/number format/precision=3,
	% 	/pgf/number format/fixed
	% },
	ymin=0,
	legend style={nodes={scale=0.8}, at= {(0.9, 0.55)}},
	height=1.75in,
	width=2.0in,
	label style={font=\normalsize},
	title style={font=\large},
	tick label style={font=\tiny}
]

% gc graph 2
\nextgroupplot[ymax=1.05, ytick={0.0, .25, .50, .75, 1 }, ylabel={$\rho$ ~|~ SSIM}
, title={German Credit}, ]

\addplot[color=blue,mark=triangle*,blue] coordinates {
(1.0, 0.249)
(5.0, 0.575)
(10.0, 0.70)
(15.0, 0.77)
(20.0, 0.81)
};
% 	(1.0, 0.279)
% (5.0, 0.538)
% (10.0, 0.647)
% (15.0, 0.706)
% (20.0, 0.741)
% };

\addplot[color=green,mark=triangle*,green] coordinates {
(1.0, 0.59)
(5.0, 0.858)
(10.0, 0.913)
(15.0, 0.938)
(20, 0.946)
};
% (1.0, 0.489)
% (5.0, 0.736)
% (10.0, 0.805)
% (15.0, 0.843)
% (20, 0.859)
% };
\addplot[color=blue,mark=triangle*,black] coordinates {
	(1.0, 0.29)

(10.0, 0.809)
(15.0, 0.863)
(20.0, 0.89)
};
% 	(1.0, 0.29)

% (10.0, 0.713)
% (15.0, 0.756)
% (20.0, 0.782)
% };

\addplot[color=green,mark=triangle*,red] coordinates {
(1.0, 0.29)

(10.0, 0.968)
(15.0, 0.978)
(20, 0.981)
};
(1.0, 0.29)
\nextgroupplot[ymax=1.05, ytick={0, .25, .50, .75, 1 }, xtick={1, 5, 10, 15, 20}, title={FMNIST}, ]

\addplot[color=blue,mark=triangle*,blue] coordinates {
	(1.0, 0.24)
(5.0, 0.58)
(10.0, 0.72)
(15.0, 0.79)
(20.0, 0.81)
};

\addplot[color=green,mark=triangle*,green] coordinates {
(1.0, 0.62)
(5.0, 0.88)
(10.0, 0.92)
(15.0, 0.95)
(20.0, 0.962)
};

\addplot[color=blue,mark=triangle*,black] coordinates {
%	(1.0, 0.24)

(10.0, 0.735)
(15.0, 0.801)
(20.0, 0.82)
};

\addplot[color=green,mark=triangle*,red] coordinates {
%(1.0, 0.62)

(10.0, 0.933)
(15.0, 0.953)
(20.0, 0.964)
};

% \addplot[color=red,mark=triangle*,red] coordinates {
% 	(0.0, 0.0)
% (0.1, 0.0)
% (0.2, 0.0)
% (0.30000000000000004, 0.0)
% (0.4, 0.0)
% (0.5, 0.0)
% (0.6000000000000001, 0.0)
% (0.7000000000000001, 0.0)
% (0.8, 0.0)
% (0.9, 0.0)

% };

% LFW GOES HERE
\nextgroupplot[ymax=1.05, ymin=0.5, ytick={ .50, .75, 1 }, title={Taiwanese},  ]
%deep base
\addplot[color=blue,mark=triangle*,blue] coordinates {
	(1.0, 0.74)
(5.0, 0.911)
(10.0, 0.945)
(15.0, 0.958)
(20.0, 0.965)
};
% 	(1.0, 0.75)
% (5.0, 0.84)
% (10.0, 0.88)
% (15.0, 0.901)
% (20.0, 0.91)
% };
\addplot[color=green,mark=triangle*,green] coordinates {
(1.0, 0.911)
(5.0, 0.95)
(10.0, 0.984)
(15.0, 0.987)
(20.0, 0.989)
};
% (1.0, 0.911)
% (5.0, 0.95)
% (10.0, 0.961)
% (15.0, 0.968)
% (20.0, 0.971)
% };

\addplot[color=blue,mark=triangle*,black] coordinates {
	%(1.0, 0.75)
(10.0, 0.963)
(15.0, 0.9727)
(20.0, 0.978)
};
% 	(1.0, 0.75)
% (10.0, 0.896)
% (15.0, 0.9105)
% (20.0, 0.918)
% };

\addplot[color=green,mark=triangle*,red] coordinates {
%(1.0, 0.911)
(10.0, 0.992)
(15.0, 0.994)
(20.0, 0.995)
};

\addlegendentry{RS}
\addlegendentry{LOO}
\addlegendentry{RS (Sel)}
\addlegendentry{LOO (Sel)}

\end{groupplot}

%\legend{Deep, Linear, Adv., Smooth};

\end{tikzpicture}
%\end{document}
}
\caption{}
\label{expl_stability}
\end{subfigure}
 \caption{Figure \subref{no_abst}: Percentage of test data with non-zero disagreement rate in normal (i.e., not selective) ensembles. Horizontal axis depicts ensemble size. While ensembling alone mitigates much of the prediction instability, it is unable to eliminate it as selective ensembles do. Figure \subref{expl_stability}: Average Spearman's Ranking coefficient, $\rho$, (For FMNIST, SSIM) between feature attributions for saliency maps generated for each individual test point (y-axis) over number of ensemble models (x-axis). The lines indicated with (Sel) in the legend are the same metrics for selective ensembles.}

\end{figure}

%While we show in Section ~\ref{ensembles_theory} that 
In this section, we demonstrate empirically that selective ensembles reduce instability in deep model predictions far below their theoretical bounds---to \emph{zero} inconsistent predictions in the test set over 276 pairwise comparisons of model predictions for each of tabular datasets, and 40 for image datasets. Additionally, following Theorem ~\ref{prop_grad_inst}, we show that feature attributions of individual deep models are frequently inconsistent, and that ensembling effectively mitigates this problem.

\paragraph{Setup.}
To evaluate selective ensembling, we focus on two sources of randomness in the learning rule: \emph{(1)} random initialization, and \emph{(2)} leave-one-out changes to the training set.
Our experiments consider seven datasets: UCI German Credit, Adult, Taiwanese Credit Default, Seizure, all from \citet{uci}; the IWPC Warfarin Dosing Recommendation~\citep{nejm-warfarin}, Fashion MNIST~\citep{xiao2017/online}, and Colorectal Histology~\citep{kather2016multi}. All of these datasets are either related to finance, credit approval, or medical diagnosis, except for FMNIST, which we include as it is a common benchmark for image classification. 
Further details %about these datasets 
are in Appendix~\ref{appendix:datasets}. 

All experiments are implemented in TensorFlow 2.3.
For each tabular, we train $500$ models from independent samples of the relevant source of randomness (e.g. leave-one-out data variations or random seeds), and for each image dataset, we train $200$ models from independent samples of each source of randomness. 
Details about the model architecture and hyperparameters used are given in Appendix~\ref{appendix:hyperparameters}.
Table~\ref{accs} reports the mean accuracy for each dataset, along with the standard deviation.

For each non-image dataset we generate 24 random ensembles of size $n \in \{5,10,15,20\}$ by selecting uniformly without replacement among the 500 pre-trained models, as well 24 ``singleton'' models drawn uniformly from the 500 to use as a point of comparison when measuring the stability of each ensemble. For image datasets, we generate 10 random ensembles of each size among 200 pre-trained models. We report ensemble predictions in the main paper using $\alpha=0.05$. %, and include results for $\alpha=0.01$ in Appendix~\ref{appendix:additional-results} as well.

\subsection{Selective Ensembles: Prediction Stability and Accuracy}
\label{sec:pred_instability}

To measure prediction instability over either selective ensembles or singleton models, we compare the predictions of each pair of models on each point in the test set, amounting to 276 comparisons for tabular datasets, and 40 comparisons for image datasets, in total for each point, and record the rate of disagreement, \pflip, across these comparisons. We report mean and variance of this disagreement over 10 random re-samplings of constituent models to create ensemble models.

The results in Table~\ref{selective_ens_flipping} and Figure~\ref{no_abst} show the percentage of points with disagreement rate greater than zero.
We see that for singleton models, as many as 57\% of test points have $\pflip > 0$, indicating that disagreement in prediction is in some cases the norm rather than the exception, although more commonly this occurs on 5-10\% of the data.
\emph{Notably, selective ensembles completely mitigate this effect:} even when \emph{as few as ten} models are included in the ensemble, \emph{no} points experienced $\pflip > 0$. %Disagreement rates of non-selective ensembles show that while ensembling on its own does help, it cannot 
Combined with the fact that abstention rates remain low (1-5\%) in all cases except where \pflip was originally very high (e.g., German Credit), this shows that selective ensembling can be a practical method for mitigating prediction instability.

Table~\ref{ens_accs_abs} shows the accuracy of selective ensembles, with abstention counted towards error, as well as accuracy of non-selective ensembles for comparison.
Notably, in all six models, with the exception of German Credit, the abstention rate drops to below $4\%$ with 20 models in the ensemble. %even with a stringent confidence threshold of $\alpha=0.01$. 
Accordingly, the accuracy of the selective ensembles in these cases is comparable---typically within a few points---to that of the traditional ensemble. However, with just five models in the ensemble, the abstention rate is 100\%; to achieve reasonable predictions with very few models, the threshold $\alpha$ needs to be increased accordingly. Disagreement of non-selective ensembles are included in Appendix~\ref{appendix:more_ens_results}: while they do lower prediction inconsistency, they do not mitigate it as effectively as selective ensembles.

\subsection{Attribution Stability}
\label{sec:grad_instability}

Following up on the theoretical result given in Proposition~\ref{prop_grad_inst}, we demonstrate that feature attributions, which are usually computed for deep models using gradients~\citep{simonyan2014deep,sundararajan2017axiomatic,leino18influence}, are often inconsistent between similar models. 
We then show that, just as ensembling increases prediction stability, it also mitigates gradient instability, leading to more consistent attributions across models.
For these experiments, we computed attributions using saliency maps~\citep{simonyan2014deep}, which are simply the gradient of the model's prediction with respect to its input, as a simple and widely-used representative of gradient-based attribution methods. 
% \begin{wrapfigure}{R}{0.6\textwidth}
% \begin{minipage}{0.6\textwidth}
\begin{figure}[t]
\resizebox{\textwidth}{!}{%
  \centering
    \includegraphics{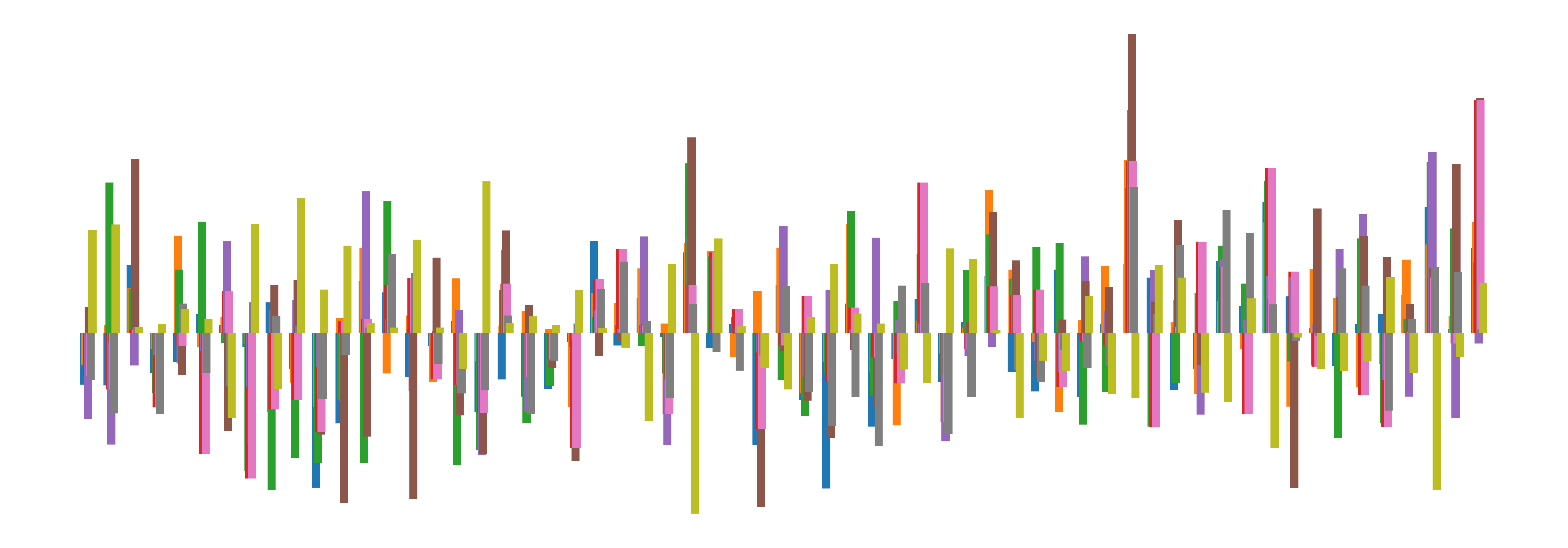}
  %   % \includegraphics[width=\textwidth]{pics/test_gc_pic2.pdf}
  % \end{minipage}
  % \begin{minipage}[b]{0.4\textwidth}
    \includegraphics{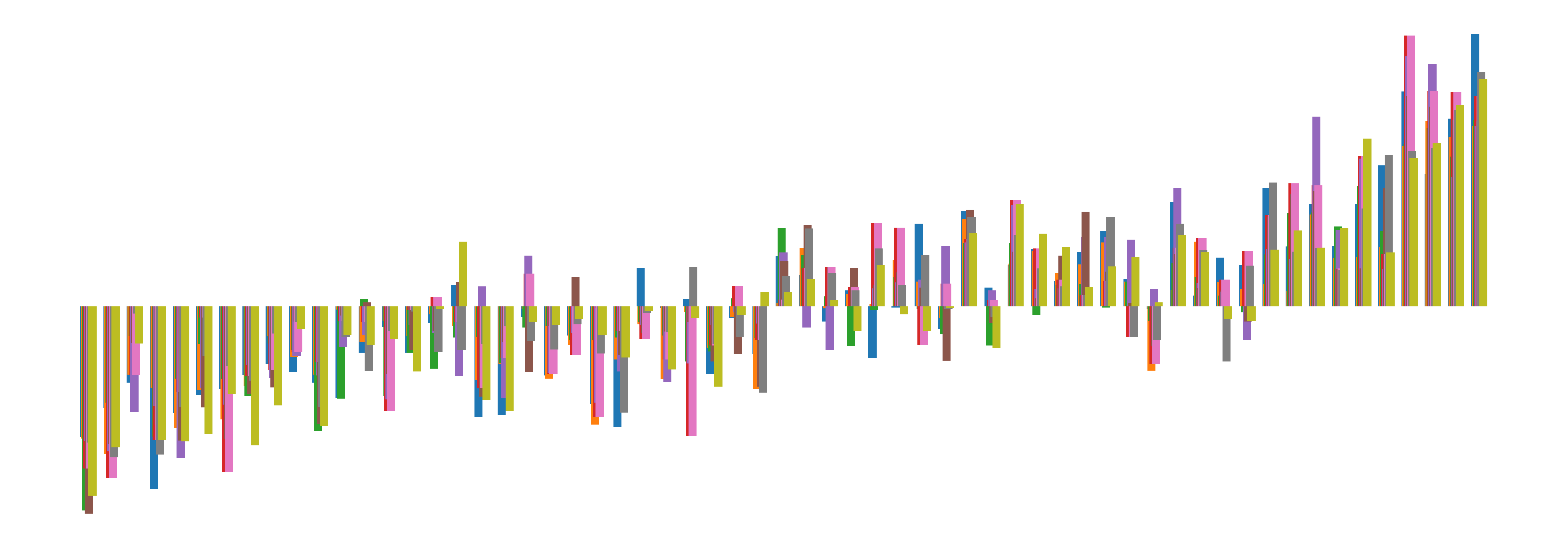}
 }
    \caption{Inconsistency of attributions on the same point across an individual (left) and ensembled (right) model ($n=15$). 
    The height of each bar on the horizontal axis represents the attribution score of a distinct feature, and each color represents a different model.
    Features are ordered according to the attribution scores of one randomly-selected model.} 
    %Note how inconsistent the individual model is, and how much more ensembles of 15 agree with each other.}
    \label{expl_inst_pic}
\end{figure}
% \end{minipage}
% \end{wrapfigure}
\paragraph{Metrics.} Following previous work~\citep{dombrowski2019explanations,ghorbani2019interpretation}, we measure the similarity between attributions using Spearman's Ranking Correlation ($\rho$) and the top-$k$ intersection, with $k=5$.
For image datasets, we also display the Structural similarity metric (SSIM), discussed further in Appendix~\ref{appendix-ssim}. Spearman's $\rho$ is a natural choice of metric as attributions induce an order of importance among features. 
We note that the top-$k$ intersection is especially interesting in tabular datasets, as often  only the most important features are of explanatory interest. 
To stay consistent with prior work, we also include Pearson's Correlation Coefficient ($r$). 
Note that $r$ and $\rho$ vary from -1 to 1, denoting negative, zero, and positive correlation. 
We compute these metrics over 276 pairwise comparisons of attributions for each size of ensemble (1, 5, 10, 15, and 20) for tabular datasets, and 40 pairwise comparisons for image datasets. 
For the top-$k$ metric, we report the mean size of the intersection between each pair of attributions.
More details are in Appendix~\ref{appendix-attr-metrics}. 

\paragraph{Baselines.}  
To contextualize the difference of attributions across models trained from distinct randomness, we also include the attribution similarity between 24 randomly chosen points in the \emph{same} model (Table~\ref{indiv_saliency}). We also present a visual comparison of model attributions, for which we simply plot the attribution for the predicted class for a given point from nine randomly selected models out of the 24, and present the feature attributions in order of their magnitude according to another randomly selected model (Figure~\ref{expl_inst_pic}).
%Further details on experimental setup are given in Section ~\ref{eval}.

%For the model attribution comparisons, we simply plotted the attribution for a given point from nine randomly selected models out of the 24, and present the feature attribution in order of their magnitude n yet another randomly selected model. For the metrics, we generated all of these metrics for each point by comparing the saliency maps for that point across two different models, for every possible combination of models. Finally, we average the metrics over all points to generate the numbers shown. Note that the top-5 intersection is computed as the average of the number of features that intersect divided by 5. Further details are in Section ~\ref{eval}. 

\paragraph{Singleton Models}

% \label{sec:indiv_grad_inst}
% We demonstrate the differences between the saliency maps of duplicitous models the same point. We use saliency maps in particular as previous work has argued for the use of gradient-based feature attribution techniques as they are faithful to the underlying model~\citep{IG,klas_first_paper}, and saliency maps are the
% %$the fundamental building block 
% simplest and most common\citep{} form of these techniques. 
% %Saliency maps are one of the most common methods of generating feature attributions in deep models, and Specifically, we show that   
% \paragraph{Variability in Gradients.}
The left image in Figure ~\ref{expl_inst_pic} demonstrates the inconsistency of model attributions of individual German Credit models on a random point in the test set. Each bar on the x-axis represents the attributions for a feature, and each different-colored bar represents a different randomly selected model. Thus, the disagreement between the sizes of the bars of different colors shows the disagreement between models on which features should be deemed important. Notably, some of the bars on the graph depicting individual models even have different signs, which means that models disagree on whether that feature counts towards or against the same prediction. 
Similar graphs for all other datasets are included in Appendix~\ref{appendix:attributions}.
%As shown in Figure~\ref{indiv_saliency}, there is significant disagreement among attributions between the models. 

We demonstrate this inconsistency further in 
%As we see in 
Table~\ref{indiv_saliency}. We see that German Credit and Seizure models have particularly unstable attributions, as the top-$k$ (and to a lesser extent, $\rho$ and $r$) scores of attributions of varying points in both the \emph{same} model, and \emph{varying models on the same point}, are quite similar. 
%We see that the saliency maps on the same point in different models are inconsistent on their most highly weighted features: for example, the 
Feature attributions of individual models are inconsistent even on highly weighted features: e.g., German Credit dataset has a top-$k$ intersection of just over one attribute on average---suggesting that attributions generated through saliency maps on these sets of models may vary substantially over benign retrainings. 
Even on models where the metrics are higher, e.g. Taiwanese Credit, the baseline similarity between attributions is higher as well---thus, we see that attributions between models of the same point are usually only 2-3$\times$ more related than those of \emph{random points within the same model}.

This instability
%e instability shown in these results 
suggests that %the 
salient variables used to inform predictions across models are %often
sensitive to random choices made during training. 
As previous work has argued in similar contexts~\citep{d2020underspecification}, this may be a result of a deep model's under-constrained search space with many local optima equivalent with respect to loss, with several %many of the 
minima corresponding to distinct \emph{rationales} for making predictions.% from observed features. 
%This perhaps points to the idea that there are so many potential mappings of inputs to outputs even within one learning rule%This is concerning from a 
%models with arbitrary changes that should be insignificant 

% In some applications, 
% %such as those with a high impact on people's lives, or those requiring close interaction with a human over time across , 
% this inconsistency may be undesirable. In the next section, we show how it is possible to lower the inconsistency between deep model predictions and decision procedures (i.e. explanations) by using selective ensembles.

\begin{table}[t]
\resizebox{\textwidth}{!}{%
\begin{tabular}{l|llll|llll}
% \toprule
        % & \multicolumn{8}{c}{Similarity Metrics on Saliency Maps (VPSM (Baseline), VMSP (Instability))} \\
        & \multicolumn{4}{c|}{Random Seed} & \multicolumn{4}{c}{Leave-one-out} \\
Dataset & Top-5 & $\rho$  & $r$ & SSIM & Top-5 & $\rho$  & $r$ & SSIM \\
\midrule
German Credit & $0.20$, $.27$ &$0.01$, $0.25$ & $0.02$, $0.28$ & -- & $0.20$, $0.49$ & $0.01$, $0.59$ & $0.02$, $0.60$  & --\\
Adult &$0.46$, $0.83$ & $0.09$, $0.83$ &$0.07$, $0.93$ & -- &$0.46$, $0.89$ & $0.15$,$0.91$ &$0.14$, $0.95$ & -- \\
Seizure & $0.14$, $0.12$ & $0.29$, $0.32$  & $0.30$, $0.33$  &  -- & $0.09$, $0.25$ & $0.23$, $0.57$  & $0.24$, $0.59$ & -- \\
Warfarin & $0.37$, $0.67 $ & $0.15$, $0.72$&  $0.12$, $0.73$ &   -- & $0.36$, $0.92 $  & $0.12$, $0.96$ & $0.11$, $0.97$  & --\\
Taiwanese Credit & $0.55$, $0.76$ & $0.35$, $0.75$  &  $0.36$,$0.83$ & --& $0.56$,$0.91$ & $0.35$,$0.95$  &  $0.37$,$0.96$  & -- \\
FMNIST & $0.00$, $0.26$ & --  & $0.61$, $0.61$  &$0.50$, $0.25$ & $0.00$, $0.57$ & --  & $0.90$, $0.89$ & $0.78$, $0.62$ \\
Colon & $0.00$, $0.63$ & --  & $0.00$, $0.92$ & $0.18$, $0.82$ & $0.00$, $0.61$ & --  & $0.00$, $0.91$& $0.18$,$0.81$ \\ 
\bottomrule
\end{tabular}}
\caption{Average top-5 intersection, Spearman's Rank Correlation Coefficient ($\rho$), and Pearson's Correlation Coefficient ($r$) to demonstrate attribution inconsistency on the \emph{same} test points across \emph{different} models. As a baseline, we compare against differences observed on \emph{different} points in the \emph{same} model. The baseline numbers are presented as: similarity baseline, similarity across models. %Intuitively, the attributions for random points in the training set should be relatively dissimilar, whereas attributions for the same point should be nearly identical--- thus, if these two numbers are close, it points to a large difference between model attributions for the same point. 
For image models, we also report the Structural Similarity Index (SSIM). Standard deviations are included in Appendix~\ref{appendix:attributions}. }
\label{indiv_saliency}
\end{table}

% \begin{figure}[t]
% \resizebox{\textwidth}{!}{%
% \input{figures/ensemble_expl_const}
% }
%  \caption{Average Spearman's Ranking coefficient, $\rho$, between feature attributions for saliency maps generated for each individual test point (y-axis) over increasing numbers of ensemble models (x-axis). %Top-5 overlap is computed by the number of overlapping features in the top five attributions, divided by five.
%   }
%     \label{expl_stability}
% \end{figure}

\paragraph{Ensemble Models.}

We demonstrate that the similarity between saliency maps of ensembled models is greater than that of individual models, and that this similarity increases linearly with  the number of models in the ensemble. For these experiments, we average each model's attributions toward the \emph{majority} predicted class of the ensemble.
 %To the best of our knowledge, there are few instances of prior work that generate feature attributions for ensemble models, and none that do so to specifically increase consistency over sources of arbitrariness in the generation process. 
%Prior works~\citep{} soliciting explanations from ensembles often only include explanations from a subset of models, as we are
On the right side of Figure ~\ref{expl_inst_pic}, we see the feature attributions of various ensemble models of size 15 over the German Credit dataset. Note how the attributions of ensemble models are much more consistent than on the individual model.

We show this phenomenon more broadly in Figure ~\ref{expl_stability}, where we display graphs of average Spearman's Rank Coefficient ($\rho$) (y-axis) between saliency maps on a point in the test set. We see $\rho$ increase as we increase the number of models in the ensemble (x-axis), for models generated over different random initializations and one-point differences in the training set. Selective ensembles can %help
further increase stability of explanations by abstaining from unstable points, and this has a marked effect when the abstention rate is high (e.g. German Credit). Similar graphs for the rest of metrics calculated %top-$k$, Pearson's Coefficient, Spearman's Coefficient, L2 distance, and SSIM for image models 
are presented in Appendix~\ref{appendix:explanation_stability}. %Again, we see a sharp increase in stability in most datasets. % by only including five models in the ensemble.

\section{Related Work}
\label{related}
%!TEX root=paper.tex

%In this paper, we mitigate the problem of inconsistent model behavior across inconsequential changes in the training environment by introducing \emph{selective ensembling}, a modeling approach that provides bounded prediction inconsistency over randomness in the training pipeline.

%In this paper, we draw attention to the problem of inconsistency in model behavior %of predictions and gradient-based explanations of deep models with 
%across inconsequential changes, such as different random initializations. 

Prior work has shown that deep models are inconsistent in their predictions across arbitrary random changes in their training pipeline, such as initialization parameters and makeup of the training set~\citep{blackleave2021,mehrer2020individual,d2020underspecification,kolen91,feldman2019shorttale}. %More recent work has demonstrated that this occurs even in duplicitous models, i.e. models with very similar aggregate accuracy~\citep{blackleave2021,mehrer2020individual}.
%In this paper, we first that this variability extends to model gradients as well, suggesting the the method by which a model makes its decision can vary between nearby models. 
%Learning rules that promote stability have received much attention in the literature to promote 
The problem of model sensitivity, particularly to variability in the training set,
%model instability more broadly (i.e. , 
%This sensitivity, specifically with respect to the training data,
can lead to an increase generalization error~\citep{elisseeff2003leave} as well as to leaking training set information~\citep{dwork2006differential,yeom2018privacy}. 
Thus, stability-enhancing learning rules have received significant attention in order to bolster desirable properties, such as privacy~\citep{liu2020intrinsic,papernot2018scalable,wang2016learningwithdp}.

One such approach is model ensembling, which has been used as a variance reduction method since the advent of statistical learning~\citep{zhou2002ensembling,valentini2004cancer,opitz1999popular,tumer1996error,dvornik2019diversity,hasan2020diabetes,freund1997decision,sagi2018ensemble,polikar2012ensemble,che2011decision,perrone1992networks,hansen1990neural}. However, to our knowledge, there is little work on providing guarantees about model disagreement using ensemble models that may \emph{abstain} from prediction. We relate our approach to the classic bias-variance decomposition of error~\citep{domingos2000unified}, showing that it certifiably bounds the variance component.

%The goal of this work is to devise a modeling process which can provide consistent predictions% in order to prevent problems resulting from prediction inconsistency in model deployment. 
%However,
Selective ensembles can be seen as a way to flag points that prone to inconsistency.
%upon which a certain model and training pipeline may give inconsistent, and thus low-quality, predictions. 
Under this view, calibration and uncertainty estimation of deep model predictions~\citep{lakshminarayanan2016simple,ovadia2019can} is a related stream of work, and one could potentially use these techniques to determine %an uncertainty threshold below which a model could abstain from prediction. 
when to abstain from prediction.
However, preventing inconsistent predictions and abstaining from uncertain predictions are different goals: in our setting, the aim is to predict the mode across models drawn from a certain distribution, whereas calibration is measured against predicting the true label.
Moreover, prior work has shown that confidence scores may not be correlated with prediction consistency across models with different random initializations~\citep{blackleave2021}. Finally, while abstaining on points with low confidence scores may lead to greater consistency, it may not yield a guarantee, which this work provides.

%While it is feasible that predicting only on points above a certain well-calibrated confidence would lead to more consistent predictions, such a method may not lead to a strong probabilistic guarantee which this work provides. Further, it is unclear how well uncertainty estimates would work to prevent inconsistency in prediction across nearby models studied in this work, as prior work has shown that confidence is not highly correlated to prediction consistency across models with changes such as different random initializations~\citep{blackleave2021}.

Conformal inference~\citep{linusson2020efficient,gupta2019nested,lofstrom2013effective}, which %aims to 
rigorously assigns confidence to predictions in settings where the data may differ from training, %be distributed differently than in training, 
is similarly related in that such a measure
%a nonconformity measure 
could be useful in identifying inconsistently predicted points. %points prone to inconsistent predictions.
%points that are prone to unstable prediction across models. 
However, in this work, we aim to achieve consistent predictions across a \emph{known} distribution of models, as prior work, as well as our results, suggest, even points conforming to past observations may still be predicted differently by different models.
%However, as prior work as well as our results suggest, even points conforming to past observations may still be predicted differently by different models. Thus, in this work we aim to achieve consistent predictions across a \emph{known} distribution of models, separating the goals of our work and those of Conformal Influence.

In addition to inconsistent predictions, this work demonstrates how feature attributions can differ substantially between individual deep models with inconsequential differences. Prior works investigating instability of gradient-based explanation techniques focus on %demonstrating instability in 
an \emph{adversarial} context~\citep{dombrowski2019explanations,ghorbani2019interpretation,heo2019fooling,wang20explanations}. %Most of these works focus on perturbing an input imperceptibly to get a different explanation, with the notable exception of 
For example, \citet{anders2020fairwashing} develop attacks to create similar models that have differing gradient-based explanations. Contrastingly, this work focuses on the instability of counterfactual explanations between similar models that may occur naturally. %, outside of an adversarial context.
As we demonstrate in Section ~\ref{sec:grad_instability}, model gradients can be quite dissimilar without any adversary.

\section{Conclusion}
\label{sec:conclusion}

We show %that in many cases, 
that similar deep models can have not only inconsistent predictions, but substantially different gradients as well.
%their gradients often differ substantially as well. 
We introduce \emph{selective ensembles} to mitigate this problem by bounding a model's inconsistency over random choices made during training. Empirically, we show that selective ensembles predict \emph{all} points consistently over all datasets we studied. 
%In cases where abstention is limited, 
Selective ensembling may present a more reliable way of using deep models in settings where high model complexity \emph{and} stability are required.%, but stability is as well.

\section*{Acknowledgments}
This work was developed with the support of NSF grant CNS-1704845,  NSF CNS-1943016, as well as by DARPA and the Air Force Research Laboratory under agreement number FA8750-15-2-0277. The U.S. Government is authorized to reproduce and distribute reprints for Governmental purposes not with- standing any copyright
notation thereon. The views, opinions, and/or findings expressed are those of the author(s) and should not be interpreted as representing the official views or policies of DARPA, the Air Force Research Laboratory, the National Science Foundation, or the U.S. Government.

\bibliography{bib}
\bibliographystyle{plainnat}

\clearpage
\newpage 
% \end{table}

\appendix

\section{Proofs}
\label{appendix:proofs}

\subsection{Proof of Theorem 3.1}
\label{app:saliency_proof}
%!TEX root=dixapp.tex

% \begin{proposition}
% Consider a binary classification model $H : R^n \rightarrow \{0,1\} = \text{sign}(h)$, where $h: R^n \rightarrow R$. Given an arbitrary function $g: R^n \rightarrow R$ which is bounded from above and below, continuous, and piecewise differentiable, and a constant $\epsilon < 0$, it is possible to construct a model $\hat{H} : R^n \rightarrow \{0,1\} = \text{sign}(\hat{h})$, $\hat{h}: R^n \rightarrow R$, where the gradients of $\hat{h}$ are determined by this arbitrary function $g$, except within $\frac{\epsilon}{2}$ of a decision boundary, but where $H(x)=\hat{H}(x) \forall x$. That is, the thresholded classification behavior is the same for both models $H$ and $\hat{H}$, but their underlying functions $h$ and $\hat{h}$ have arbitrarily different gradients, and therefore arbitrarily different gradient-based explanations, almost everywhere. %maintains the same mapping $\hat{H}: R^n \rightarrow \{0,1\} = h: S \rightarrow \{0,1\}$.
% %Specifically, this means that $h$ and $\hat{h}$ will have the same 0/1 loss and prediction behavior as $h$, but arbitrarily different gradient-based explanations. 
% \end{proposition}

\begin{figure*}[!tbp]
\resizebox{\textwidth}{!}{%
  \centering
  \begin{minipage}[b]{0.5\textwidth}
    \includegraphics[width=\textwidth]{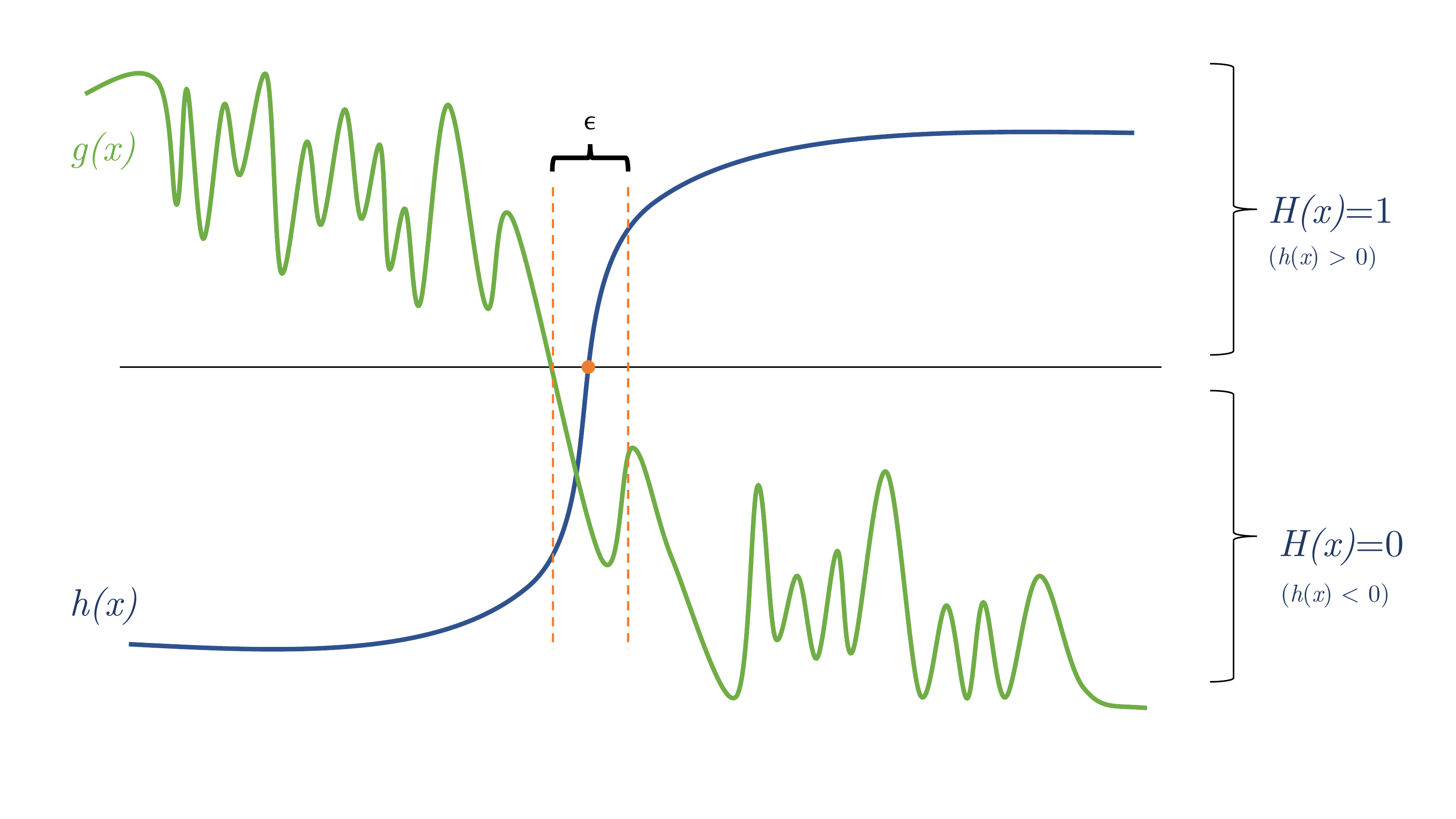}
  \end{minipage}
  \hfill
  \begin{minipage}[b]{0.5\textwidth}
    \includegraphics[width=\textwidth]{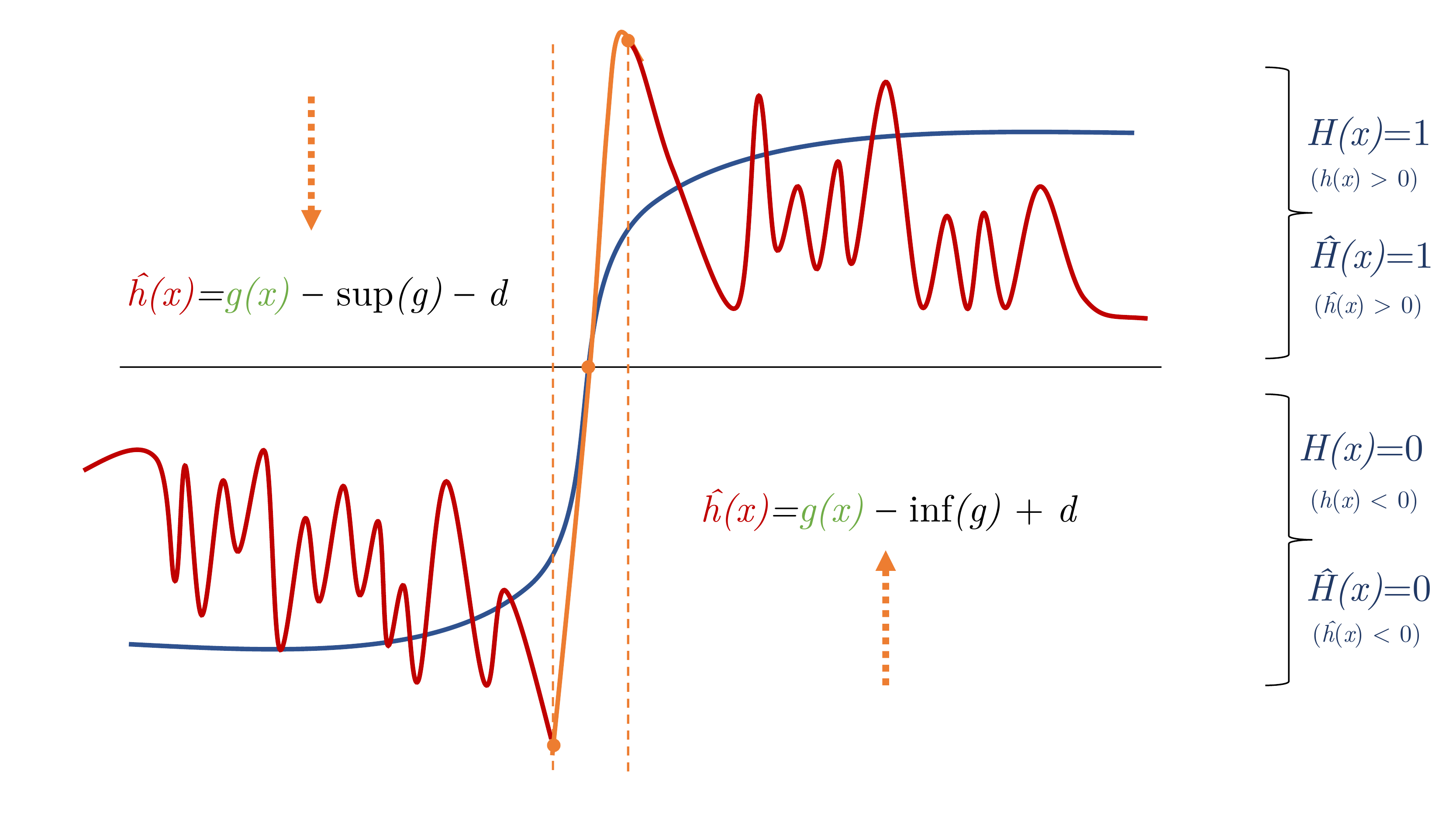}
  \end{minipage}
 }
 \caption{Intuitive illustration of how two models which predict identical classification labels can have arbitrary gradients. To show this, given a binary classifier $H$ and an arbitrary function $g$, we construct a classifier $H'$ that predicts the same labels as $H$, yet has gradients equal to $g$ almost everywhere. We formally state this result in Theorem~\ref{prop_grad_inst}.}
 \label{app:expl_proof_int}
\end{figure*}

\paragraph{Theorem 3.1.}\hspace{-0.5em}\textit{
\label{prop_grad_inst}
Let $H : \X \rightarrow \{-1,1\} = \mathrm{sign}(h)$ be a binary classifier and
$g: \X \rightarrow \mathbb{R}$ be an unrelated function that is bounded from above and below, continuous, and piecewise differentiable.
Then there exists another binary classifier $\hat H = \mathrm{sign}(\hat h)$ such that for any $\epsilon > 0$,
$$
\forall x\in\X~.\hspace{2em}
\text{1.}~~\hat H(x) = H(x)\hspace{2em}
\text{2.}~~\inf_{x' : H(x')\neq H(x)}\Big\{||x - x'||\Big\} > \nicefrac{\epsilon}{2}~~\Longrightarrow~~\nabla \hat h(x) = \nabla g(x)$$
}
\begin{proof}
We partition $\X$ into regions $\{I_1....I_k\}$ determined by the decision boundaries of $H$. That is, each $I_i$ represents a maximal contiguous region for which each $x\in I_i$ receives the same label from $H$. 

%We are given a function $g$, which we divide into functions $g_i(x): x \in I_i \rightarrow R$, i.e. we divide the range of $g$ to create functions that are specific to each decision region $\{I_1....I_k\}$ of $H$. 

%Then, we create a set of functions $\hat{g_{I_i}}(x): x \in I_i \rightarrow R$ such that
Recall we are given a function $g: \X \rightarrow \mathbb{R}$ which is bounded from above and below. We create a set of functions $\hat{g_{I_i}}: I_i \rightarrow R$ such that 
\[
\hat{g}_{I_i}(x)= \begin{cases} 
g(x)-\inf_x g(x) + c  &\text{    if $H(I_i)=1$} \\
g(x)-\sup_x g(x) - c &\text{    if $H(I_i)=-1$}
\end{cases}
\]%Consider the function $\hat{g}= g_i(x)$ for $x \in I_i$. 

where $c$ is some small constant greater than zero. Additionally, let $d(x)$ be the $\ell_2$ distance from $x$ to the nearest decision boundary of $h$, i.e. $d(x)=\inf_{x' : H(x')\neq H(x)}\Big\{||x - x'||\Big\}$.  Then, we define $\hat{h}$ to be:

\[ \hat{h}(x) =  \begin{cases}
\hat{g}_{I_i}(x) &\text{ for $x$ } \in I_i \text{ if $d(x) > \frac{\epsilon}{2}$}\\
\hat{g}_{I_i}(x) \cdot \frac{2d(x)}{\epsilon} &\text{ for $x$ } \in I_i \text{ if $d(x) \leq \frac{\epsilon}{2}$} \\
\end{cases}
\]

And, as described above, we define $\hat{H}=\text{sign}(\hat{h})$. 
First, we show that $\hat{H}(x)=H(x)~~\forall x \in \X$.
Without loss of generality, consider some $I_i$ where $H(x)=1$, for any $x \in I_i$. We first consider the case where $d(x)> \frac{\epsilon}{2}$. 

By construction, for $x \in I_i$, $\hat{H}(x)= \text{sign}(\hat{h}(x))= \text{sign}(\hat{g}_{I_i}(x)) = \text{sign}(g(x)-\inf_x g(x)+ c)$.
By definition of the infimum, $g(x)-\inf_x g(x)\geq 0$, and thus $\text{sign}(g(x)-\inf_x g(x)+ c)=1$, so $\hat{H}(x)=1=H(x)$.

Note that in the case where $d(x) \leq \frac{\epsilon}{2}$, we can follow the same argument as multiplication by a positive constant does not affect the sign.
A symmetric argument follows for the case where for $x\in I_i$, $H(x)=-1$; thus, $\hat{H}(x)=H(x)~~\forall x \in \X$.

Secondly, we show that $\nabla \hat{h}(x)= \nabla g(x)~~\forall x$ where $d(x) > \frac{\epsilon}{2}$. Consider the case where $H(x)=1$.
By construction, $\hat{h}(x)=\hat{g}_{I_i}(x)= g(x)-\inf_x g(x) + c$. Note that this means the infimum and $c$ are constants, so their gradients are zero. Thus, $\nabla \hat{h}(x)= \nabla g(x)$. A symmetric argument holds for the case where $H(x)=-1$.

%Note that this construction necessitates that $\hat{h}(x)=0$ if and only if $h(x)=0$, as $g_i(x)-\inf_x g_i(x)>0$, and $\hat{g}(x)$ is defined this way when $h(x)=0$. Similarly, $\hat{g}(x)=1$ if and only if $h(x)=1$. Further, see that $\nabla \hat{g}(x)= \nabla g(x) \forall x \text{where} d(x) < \epsilon$, as $\hat{g_i}(x)$ and $g_i(x)$ only differ by a constant for all $i$, and thus the gradient must be the same.   

It remains to prove that $\hat{h}$ is continuous and piecewise differentiable, in order to be a realizable as a ReLU-network. By assumption, $g$ is piecewise differentiable, which means that $\hat{g_i}$ are piecewise differentiable as well, as is $\hat{g_i}(x) \cdot \frac{d(x)}{\epsilon}$. Thus, $\hat{h}$ is piecewise-differentiable. To see that $\hat{h}$ is continuous, consider the case where $d(x)=\nicefrac{\epsilon}{2}$ for some $x$. Then $\hat{g_i}(x) \cdot \frac{d(x)}{\epsilon} = \hat{g_i}(x) \cdot \frac{\epsilon}{\epsilon}= \hat{g_i}(x)$. Additionally, consider the case where $d(x)=0$, i.e. $x$ is on a decision boundary of $h(x)$, between two regions $I_i, I_j$. Then $\hat{h}(x)=\hat{g_i}(x) \cdot \frac{d(x)}{\epsilon}= \hat{g_i}(x) \cdot 0= 0= \hat{g_j}(x) \cdot 0= \hat{g_j}(x)$. This shows that the piecewise components of $\hat{h}$ come to the same value at their intersection.%, and so $\hat{h}$ is continuous. 
Further, each piecewise component of $\hat{h}$ is equal to some continuous function, as $g(x)$ is continuous by assumption. Thus, $\hat{h}$ is continuous, and we conclude our proof.
%when $d(x) \neq 0, \epsilon$, then $\hat{h}$ is continuous since $g$ is continuous as %required, and thus $\hat{g_i}$ are continuous $\forall i$, as well as $\hat{g_i}\cdot \frac{2d(x)}{\epsilon}$, as the multiplied expression is simply a constant.    
\end{proof}

We include a visual intuition of the proof in Figure~\ref{app:expl_proof_int}.
% \begin{proposition}
% Given a binary classification model $h: R^n \rightarrow \{0,1\}$, and a constant $\epsilon < 0$, it is possible to construct a model $\hat{h}$ with arbitrary gradients everywhere except within $\epsilon$ of a decision boundary, but maintains the same mapping $\hat{h}: R^n \rightarrow \{0,1\} = h: R^n \rightarrow \{0,1\}$. Specifically, this means that $h$ and $\hat{h}$ will have the same 0/1 loss, but arbitrarily different gradient-based explanations. 
% \end{proposition}
% \begin{proof}
% We divide the 
% Suppose we are given $g$

% \end{proof}

% \begin{proposition}
% Consider a binary classfication model $h: R^n \rightarrow [0,1]$, whose outputs we threshold to determine a mapping $h: R^n \rightarrow \{0,1\}$. We can create a model $\hat{h}$ with arbitrary explanations, but maintains the same mapping from 
% \end{proposition}
% \begin{proof}

% \end{proof}

\subsection{Proof of Theorem 4.1}
\label{app:rand_smooth_proof}
%!TEX root=dixapp.tex

\paragraph{Theorem 4.1.}\hspace{-0.5em}\textit{
\label{thm:matches_mode_app}
Let $\pipeline$ be a learning pipeline, and let $\randomness$ be a distribution over random states.
Further, let $g_{\pipeline,\randomness}$ be the mode predictor, let $\hat{g}_n(\pipeline,S)$ for $S\sim\randomness^n$ be a selective ensemble, and let $\alpha \geq 0$.
Then,
$$
\forall x\in\X~~.~~
\pr_{S\sim\randomness^n}\Big[
\hat{g}_n(\pipeline,S~;~\alpha, x) \neqorabs g_{\pipeline,\randomness}(x)
\Big] \leq \alpha
$$ 
}
\begin{proof}
$\hat{g}_n(\pipeline,S)$ is an ensemble of $n$ models. By the definition of Algorithm~\ref{prediction_algo},  $\hat{g}_n(\pipeline,S)$ gathers a vector of class counts of the prediction for $x$ from each model in the ensemble. Let the class with the highest count be $c_A$, with counts $n_A$, and the class with the second highest count be called $c_B$, with counts $n_B$. $\hat{g}_n(\pipeline,S)$ runs a two-sided hypothesis test to ensure that $\Pr[n_A \sim \text{Binomial}(n_A+n_B, 0.5)] < \alpha$, i.e. that $c_A$ is the true mode prediction over $\randomness$. See that
\begin{align}
&\Pr\Big[\modepredictor(x) \neq c_A ~~\land~~ \ensembleat{x}=c_A \Big] \\ 
=~&\Pr\Big[\modepredictor(x) \neq c_A\Big] \cdot \Pr\Big[\ensembleat{x} \neq \mathtt{ABSTAIN}~|~\modepredictor(x) \neq c_A\Big] \\
\leq~&\Pr\Big[\ensembleat{x}\neq \mathtt{ABSTAIN} ~|~ \modepredictor(x) \neq c_A\Big] \\
    	% \text{By Hung and Fithian\citep{}, we have that}
\leq~&\Pr\Big[\ensembleat{x} \neq \mathtt{ABSTAIN} ~|~ \modepredictor(x) \neq c_A\Big] = \alpha & \text{By \citet{hung2019rank}}
\end{align}
Thus,
$$\Pr\Big[\modepredictor(x) \neq c_A ~~\land~~ \ensembleat{x}=c_A \Big] \leq \alpha$$
\end{proof}

\subsection{Proof of Corollary 4.2}
\label{app:corollary4.2}
%!TEX root=dixapp.tex

\paragraph{Corollary 4.2.}\hspace{-0.5em}\textit{
\label{corr:loss_variance}
Let $\pipeline$ be a learning pipeline, and let $\randomness$ be a distribution over random states.
Further, let $g_{\pipeline,\randomness}$ be the mode predictor, let $\hat{g}_n(\pipeline,S)$ for $S\sim\randomness^n$ be a selective ensemble.
Finally, let $\alpha \geq 0$, and let $\beta \geq 0$ be an upper bound on the expected abstention rate of $\hat{g}_n(\pipeline,S)$.
Then, the expected \emph{loss variance}, $V(x)$, over inputs, $x$, is bounded by $\alpha + \beta$.
That is,
$$
\E_{x\sim\mathcal D}\Big[V(x)\Big] 
= \E_{x\sim\mathcal D}\Bigg[~
	\Pr_{S\sim\randomness^n}\Big[ \hat{g}_n(\pipeline,S~;~x) \neq g_{\pipeline,\randomness}(x) \Big]
~\Bigg] 
\leq \alpha + \beta
$$}

\begin{proof}
Since $\modepredictor$ never abstains, we have by the law of total probability that
\begin{footnotesize}
\begin{align*}
\Pr_{S\sim\randomness^n}\Big[ \ensembleat{x} \neq \modepredictor(x) \Big]
&= \Pr_{S\sim\randomness^n}\Big[ 
	\ensembleat{x} \neqorabs \modepredictor(x) ~~\lor~~ \ensembleat{x} = \mathtt{ABSTAIN} \\
&= \Pr_{S\sim\randomness^n}\Big[ 
	\ensembleat{x} \neqorabs \modepredictor(x)\Big] + \Pr_{S\sim\randomness^n}\Big[\ensembleat{x} = \mathtt{ABSTAIN}
\Big]
\end{align*}
\end{footnotesize}
By Theorem~\ref{thm:matches_mode}, we have that $\Pr_{S\sim\randomness^n}\big[\ensembleat{x} \neqorabs \modepredictor(x)\big] \leq \alpha$, thus
$$
\E_{x\sim\mathcal D}\Bigg[~\Pr_{S\sim\randomness^n}\Big[ \ensembleat{x} \neq \modepredictor(x) \Big]~\Bigg]
\leq \alpha + \E_{x\sim\mathcal D}\Bigg[~\Pr_{S\sim\randomness^n}\Big[\ensembleat{x} = \mathtt{ABSTAIN}\Big]~\Bigg]
$$
Finally, since $\beta$ is an upper bound on the expected abstention rate of $\ensemble$, we conclude that
$$
\E_{x\sim\mathcal D}\Bigg[~\Pr_{S\sim\randomness^n}\Big[ \ensembleat{x} \neq \modepredictor(x) \Big]~\Bigg]
\leq \alpha + \beta
$$
\end{proof}

\subsection{Proof of Corollary 4.3}
\label{app:corollary4.3}
%!TEX root=dixapp.tex
\paragraph{Corollary 4.3.}\hspace{-0.5em}\textit{
\label{prop:disagreement}
Let $\pipeline$ be a learning pipeline, and let $\randomness$ be a distribution over random states.
Further, let $\hat{g}_n(\pipeline,S)$ for $S\sim\randomness^n$ be a selective ensemble.
Finally, let $\alpha \geq 0$, and let $\beta \geq 0$ be an upper bound on the expected abstention rate of $\hat{g}_n(\pipeline,S)$.
Then,
$$
\E_{x\sim\mathcal D}\Bigg[~
\pr_{S^1,S^2\sim\randomness^n}\Big[
\hat{g}_n(\pipeline,S^1~;~\alpha, x) \neq \hat{g}_n(\pipeline,S^2~;~\alpha, x)
\Big] 
~\Bigg] 
\leq 2(\alpha + \beta)
$$
}

\begin{proof}
For $i\in\{1,2\}$, let $A^i$ be the event that $\hat{g}_n(\pipeline,S^i~;~\alpha, x) = \mathtt{ABSTAIN}$, and let $N^i$ be the event that $\hat{g}_n(\pipeline,S^i~;~\alpha, x) \neqorabs \modepredictor$.
In the worst case, $A^1$ and $A^2$, and $N^1$ and $N^2$ are disjoint, that is, e.g., if $\hat{g}_n(\pipeline,S^i)$ abstains on $x$, then $\hat{g}_n(\pipeline,S^1~;~\alpha, x) \neq \hat{g}_n(\pipeline,S^2~;~\alpha, x)$.
In other words, we have that
$$
\pr_{S^1,S^2\sim\randomness^n}\Big[
\hat{g}_n(\pipeline,S^1~;~\alpha, x) \neq \hat{g}_n(\pipeline,S^2~;~\alpha, x)
\Big] \leq \pr\Big[A^1 ~\lor~ A^2 ~\lor~ N^1 ~\lor~ N^2\Big]
$$
which, by union bound implies that
$$
\pr_{S^1,S^2\sim\randomness^n}\Big[
\hat{g}_n(\pipeline,S^1~;~\alpha, x) \neq \hat{g}_n(\pipeline,S^2~;~\alpha, x)
\Big] \leq \pr\big[A^1\big] + \pr\big[A^2\big] + \pr\big[N^1\big] + \pr\big[N^2\big].
$$
By Theorem~\ref{thm:matches_mode} $\Pr\big[N^i\big] \leq \alpha$.
Thus we have
$$
\E_{x\sim\mathcal D}\Bigg[~
	\pr_{S^1,S^2\sim\randomness^n}\Big[
	\hat{g}_n(\pipeline,S^1~;~\alpha, x) \neq \hat{g}_n(\pipeline,S^2~;~\alpha, x)
	\Big]
~\Bigg]
\leq 
2\alpha + 
\E_{x\sim\mathcal D}\Big[~\pr\big[A^1\big]~\Big] +
\E_{x\sim\mathcal D}\Big[~\pr\big[A^2\big]~\Big].
$$
Finally, since $\beta$ is an upper bound on the expected abstention rate of $\ensemble$, we conclude that
$$
\E_{x\sim\mathcal D}\Bigg[~
	\pr_{S^1,S^2\sim\randomness^n}\Big[
	\hat{g}_n(\pipeline,S^1~;~\alpha, x) \neq \hat{g}_n(\pipeline,S^2~;~\alpha, x)
	\Big]
~\Bigg]
\leq 
2(\alpha + \beta)
$$
\end{proof}

\section{Datasets}
\label{appendix:datasets}
The German Credit and Taiwanese data sets consist of individuals financial data, with a binary response indicating their creditworthiness. For the German Credit dataset, there are 1000 points, and 20 attributes. We one-hot encode the data to get 61 features, and standardize the data to zero mean and unit variance using SKLearn Standard scaler. We partitioned the data intro a training set of 700 and a test set of 200.  
The Taiwanese credit dataset has 30,000 instances with 24 attributes. We one-hot encode the data to get 32 features and normalize the data to be between zero and one. We partitioned the data intro a training set of 22500 and a test set of 7500.  

The Adult dataset consists of a subset of publicly-available US Census data, binary response indicating annual income of $>50$k. There are 14 attributes, which we one-hot encode to get 96 features. We normalize the numerical features to have values between $0$ and $1$. After removing instances with missing values, there are $30,162$ examples which we split into a training set of 14891, a leave one out set of 100, and a test set of 1501 examples. 

The Seizure dataset comprises time-series EEG recordings for 500 individuals, with a binary response indicating the occurrence of a seizure. This is represented as 11500 rows with 178 features each. We split this into 7,950 train points and 3,550 test points. We standardize the numeric features to zero mean and unit variance. 

The Warfain dataset is collected by the International Warfarin
Pharmacogenetics Consortium~\citep{nejm-warfarin} about patients who were
prescribed warfarin. We removed rows with missing
values, 4819 patients remained in the dataset. The inputs to the
model are demographic (age, height, weight, race), medical
(use of amiodarone, use of enzyme inducer), and genetic
(VKORC1, CYP2C9) attributes. Age, height, and weight are
real-valued and were scaled to zero mean and unit variance.
The medical attributes take binary values, and the remaining
attributes were one-hot encoded. The output is the weekly dose
of warfarin in milligrams, which we encode as "low", "medium", or "high", following the recommendations set by~\citet{nejm-warfarin}.

Fashion MNIST contains images of clothing items, with a multilabel response of 10 classes. 
There are 60000 training examples and 10000 test examples. We pre-process the data by normalizing the numerical values in the image array to be between $0$ and $1$.

The colorectal histology dataset contains images of human colorectal cancer, with a multilabel response of 8 classes. There are 5,000 images, which we divide into a training set of 3750 and a validation set of 1250. We pre-process the data by normalizing the numerical values in the image array to be between $0$ and $1$.

The UCI datasets as well as FMNIST are under an MIT license, the colorectal histology and Warfarin datasets are under a Creative Commons License.~\citep{uci,colon_license,nejm-warfarin}.

%The German Credit and Taiwanese Credit data sets consists of individuals’ financial data, with a binary response indicating their creditworthiness. The Adult dataset consists of a subset of publicly-available US Census data, with a binary response indicating annual income of > 50k. The Seizure dataset comprises time-series EEG recordings for 500 individuals, with a binary response indicating the occurrence of a seizure. The Warfarin Dosing dataset consists of indivduals health and demographic information, with warfarin dosage recommendations of "low", "medium," and "high" as labels. Fashion MNIST contains images of clothing items, with a multilabel response of 10 classes. Correctal Histology consists of  pictures of individuals’ colons, with eight response labels connoting the identity health of their colon. Further information about these datasets and the preprocessing steps we apply can be found in the supplementary material. Table ~\ref{accs} contains the mean accuracy and generalization error for the 500 models we trained on each dataset, along with the maximum deviation of accuracy from the mean. 

\section{Model Architecture and Hyper-Parameters}
\label{appendix:hyperparameters}
The German Credit and Seizure models have
three hidden layers, of size 128, 64, and 16. Models on the Adult
dataset have one hidden layer of 200 neurons. Models on the Taiwanese dataset have two hidden layers of 32 and 16. The Warfarin models have one hidden layer of 100. The FMNIST model
is a modified LeNet architecture~\citep{lecun1995learning}. This model is trained with
dropout. The Colon models are trained with a modified, ResNet50~\citep{he2016deep}, pre-trained on ImageNet~\citep{deng2009imagenet}, available from Keras.
German Credit,
Adult, Seizure, Taiwanese, and Warfarin models are trained for 100 epochs; FMNIST for 50, and Colon models are trained for 20 epochs. German Credit models are trained
with a batch size of 32; FMNIST 64; Adult, Seizure, and Warfarin models with
batch sizes of 128; and Colon and Taiwanese Credit models with batch sizes of 512. German Credit, Adult, Seizure, Taiwanese Credit, Warfarin, and Colon are trained with keras' Adam optimizer with the default parameters. FMNIST models are trained with keras' SGD optimizer with the default parameters.

Note that we discuss train-test splits and data preprocessing above in Section~\ref{appendix:datasets}. We prepare different models for the same dataset using Tensorflow 2.3.0 and all computations are done using a Titan RTX accelerator on a machine with 64 gigabytes of memory. 

\section{Metrics}
\label{appendix-attr-metrics}

We report similarity between feature attributions with Spearman's Ranking Correlation ($\rho$), Pearson's Correlation Coefficient ($r$), top-$k$ intersection, $\ell_2$ distance, and SSIM for image datasets. We use standard implementations for Spearman's Ranking Correlation ($\rho$) and Pearson's Correlation Coefficient ($r$) from scipy, and implement  $\ell_2$ distance as well as the top-$k$ using numpy functions.

Note that $r$ and $\rho$ vary from -1 to 1, denoting negative, zero, and positive correlation. We display top-$k$ for $k$=5, and compute this by taking the number of features in the intersection of the top $5$ between two models, and then diving this by $5$. Thus top-$k$ is between 0 and 1, indicating low and high correlation respectively. 

The $\ell_2$ distance has a minimum of $0$, but is unbounded from above, and SSIM varies from -1 to 1, indicating no correlation to exact correlation respectively.

Note that we compute these metrics between two different models on the same point, for every point in the test set, over 276 different pairs of models for tabular datasets and over 40 pairs of models for image datasets. We average this result over the points in the test set and over the comparisons to get the numbers displayed in the tables and graphs throughout the paper.

%To stay consistent with prior work, we also include Pearson's Correlation Coefficient ($r$). 
%Note that $r$ and $\rho$ vary from -1 to 1, denoting negative, zero, and positive correlation. 
\subsection{SSIM}
%Comparing feature attributions for image models numerically is difficult, as each numerical attribution corresponds to one pixel out of hundreds or thousands. %Additionally, images with large pixel-level differences may be very similar visually, for example two copies of the exact same image where one is shifted over by one row of pixels. 
Explanations for image models can be interpreted as an image (as there is an attribution for each pixel), and are often evaluated visually~\citep{leino18influence,simonyan2014deep,sundararajan2017axiomatic}. However, pixel-wise indicators for similarity between images (such as top-k similarity between pixel values, Spearman's ranking coefficient, or mean squared error) often do not capture how similar images are visually, in aggregate. In order to give an indication if the entire explanation for an image model, i.e. the explanatory image produced, is similar, we use the structural similarity index (SSIM)~\citep{wang2004image}. We use the implementation from $\mathtt{scikit-image}$~\citep{structural_similarity_index}. SSIM varies from -1 to 1, indicating no correlation to exact correlation respectively.
\label{appendix-ssim}

\section{Experimental Results for $\alpha=0.01$}
\label{appendix:additional-results}
We include results on the prediction of selective ensemble models for $\alpha=0.01$ as well. We include the percentage of points with disagreement between at least one pair of models ($\pflip > 0$) trained with different random seeds (RS) or leave-one-out differences in training data, for singleton models ($n=1$) and selective ensembles ($n > 1$) in Table~\ref{app:selective_ens_flipping}. Notice the number of points with $\pflip > 0$ is again zero. We also include the mean and standard deviation of accuracy and abstention rate for $\alpha=0.01$ in Table~\ref{app:tab:ens_acc_abs01}.

\begin{table}[t]
\resizebox{\textwidth}{!}{%
  \begin{tabular}{ll|lllllll}
% \multicolumn{4}{c}{portion of Taiwanese Credit test data with $\pflip > 0$} \\
%Randomness& $n$ & Mean & Std \\
\multicolumn{7}{c}{mean $\pm$ std. dev of portion of test data with $\pflip > 0$} \\
Randomness& $n$ & Ger. Credit & Adult  &  Seizure & Tai. Credit & Warfarin  & FMNIST &  Colon\\
\midrule
 %& & Mean & Std.\\
RS & 1 & $.570 \pm .020$ & $.087 \pm .001 $ & $.060 \pm .01$ &   $.082 \pm .002$ & $.098 \pm .003$ & $.061 \pm .008$  & $.037 \pm .005$   \\
RS & (5, 10, 15, 20) & $0.0 \pm 0.0$ & $0.0\pm 0.0$ & $0.0 \pm 0.0$ & $0.0 \pm 0.0$ & $0.0 \pm 0.0$ & $0.0 \pm 0.0$ & $0.0 \pm 0.0$  \\
LOO & 1 &  $.262 \pm .014$ & $.063 \pm .001$ & $.031 \pm .001$ & $.031 \pm .001$ & $.033 \pm .003$ & $.034 \pm .004$ & $.042 \pm .005$ \\
LOO &(5, 10, 15, 20) & $0.0 \pm 0.0$ & $0.0\pm 0.0$ & $0.0 \pm 0.0$ & $0.0 \pm 0.0$ & $0.0 \pm 0.0$ & $0.0 \pm 0.0$ & $0.0 \pm 0.0$ \\
\bottomrule
\end{tabular}
}
\caption{The percentage of points with disagreement between at least one pair of models ($\pflip > 0$) trained with different random seeds (RS) or leave-one-out differences in training data, for singleton models ($n=1$) and selective ensembles ($n > 1$).
Results for selective ensembles all selective ensembles are shown together, as they all have no disagreement. Note that these results are for \textbf{$\alpha=0.01$}. But this different $\alpha$ also leads to zero disagreement between predicted points.}
\label{app:selective_ens_flipping}
\end{table}

\begin{table}
\resizebox{\textwidth}{!}{%
\begin{tabular}{ll|l@{\hskip 1pt}l@{\hskip 1pt}l@{\hskip 1pt}l@{\hskip 1pt}l@{\hskip 1pt}l@{\hskip 1pt}l@{\hskip 1pt}l}
% \toprule
    &     \multicolumn{8}{c}{\emph{mean accuracy (abstain as error) / std. dev}} \\
\randomness & $n$ & Ger. Credit & Adult &  Seizure & Wafarin & Tai. Credit & FMNIST & Colon \\
\midrule
RS & 5 & $0.0\pm0.0$ & $0.0\pm0.0$ & $0.0\pm0.0$ & $0.0\pm0.0$ & $0.0\pm0.0$ & $0.0\pm0.0$ & $0.0\pm0.0$ \\
RS & 10 & $.461\pm.016$  & $.807\pm1e-3~$ & $.945\pm2e-3$ & $.646\pm3e-3$ &  $.788\pm2e-3$   & $.870\pm5e-3$ & $.902\pm2e-3$ \\
RS & 15 & $.589\pm.015$  & $.822\pm8e-4~$  & $.961\pm1e-3$ &  $.661\pm3e-3$ & $.802\pm9e-4$  & $.890\pm2e-3$ & $.915\pm1e-3$  \\
RS & 20 & $.593\pm.011$  & $.822\pm7e-4~$  & $.961\pm8e-4$ &  $.662\pm1e-3$& $.803\pm9e-4$  & $.991\pm1e-3$   & $.926\pm1e-3$ \\
\midrule
LOO & 5  & $0.0\pm0.0$ & $0.0\pm0.0$ & $0.0\pm0.0$ & $0.0\pm0.0$ & $0.0\pm0.0$ & $0.0\pm0.0$ & $0.0\pm0.0$ \\
LOO & 10  & $.618\pm.017$~~~~~~ & $.818\pm1e-3$~~~~~~ & $.947\pm4e-3$~~~~~~  & $.674\pm2e-3$~~~~~~& $.807\pm1e-3$~~~~~~ & $.904\pm6e-4$~~~~~~  & $.901\pm2e-3$~~~~~~ \\
LOO & 15  & $.656\pm.017$ &$.828\pm1e-3~$ & $.963\pm1e-3$ & $.678\pm9e-4$ & $.812\pm9e-4$  & $.908\pm1e-3$  & $.912\pm2e-3$ \\
LOO & 20  & $.661\pm.018$ & $.829\pm7e-4~$ & $.964\pm1e-3$ & $.678\pm7e-4$& $.812\pm8e-4$ & $.909\pm6e-4$ & $.912\pm2e-3$  \\
% \end{tabular}
 \\
% \begin{tabular}{ll|l@{\hskip 1pt}l@{\hskip 1pt}l@{\hskip 1pt}l@{\hskip 1pt}l@{\hskip 1pt}l@{\hskip 1pt}l@{\hskip 1pt}l}
 \toprule
         & \multicolumn{7}{c}{\emph{mean abstention rate / std dev}} \\
\randomness & $n$ & Ger. Credit & Adult &  Seizure & Warfarin & Tai. Credit & FMNIST & Colon \\
\midrule
RS & 5 & $1.0 \pm 0.0 $ & $1.0 \pm 0.0$ & $1.0 \pm 0.0$ & $1.0 \pm 0.0$ & $1.0 \pm 0.0$ & $1.0 \pm 0.0$ & $1.0 \pm 0.0$ \\
RS & 10 & $.449\pm .021$ & $ .068\pm2e-3$ & $ .045\pm2e-3$  & $.078\pm5e-3$ &  $.063\pm2e-3$ & $.087\pm8e-3$ & $.050\pm3e-3$ \\
RS & 15 & $.278\pm .017$ & $.041\pm1e-3$  & $.025\pm1e-3$ & $.049\pm3e-3$ & $.037\pm1e-3$  & $.055\pm2e-3$ & $.030\pm2e-3$ \\
RS & 20 & $.270\pm.015$ & $.040\pm1-e3$  & $.024\pm1e-3$  &  $.047\pm2e-3$ & $.036\pm1e-3$  & $.054\pm9e-4$  & $.038\pm1e-3$ \\
\midrule
LOO & 5  & $1.0 \pm 0.0 $ & $1.0 \pm 0.0$ & $1.0 \pm 0.0$ & $1.0 \pm 0.0$ & $1.0 \pm 0.0$ & $1.0 \pm 0.0$ & $1.0 \pm 0.0$ \\
LOO & 10  & $.215 \pm .030$~~~~~~ & $.049\pm2e-3$~~~~~~ & $.045\pm5e-3$~~~~~~& $.027\pm2e-3$~~~~~~& $.025\pm1e-3$~~~~~~ & $.029\pm1e-3$~~~~~~   & $.054\pm2e-3$~~~~~~ \\
LOO & 15  & $.144\pm 0.040$  &$.030\pm2e-3$ & $.026\pm1e-3$ &$.017\pm2e-3$& $.017\pm2e-3$ & $.021\pm3e-3$  & $.035\pm2e-3$ \\
LOO & 20  & $.135\pm.040$ &  $.029\pm1e-3$& $.025\pm1e-3$ & $.017\pm1e-3$& $.017\pm2e-3$ & $.019\pm1e-3$ & $.035\pm3e-3$ \\
\bottomrule
\end{tabular}
}
\caption{Accuracy (above) and abstention rate (below) of selective ensembles with $n \in \{5, 10, 15, 20\}$ constituents.
Results are averaged over 24 models, standard deviation is presented. Note that these results are for \textbf{alpha=0.01}. }
\label{app:tab:ens_acc_abs01}
\end{table}

\section{Selective Ensembling Full Results}
\label{appendix:more_ens_results}

We include the full results from the evaluation section, including error bars on the disagreement, accuracy, abstention rates of selective ensembles, in Table~\ref{app:tab:ens_disagree_std} and Table~\ref{app:tab:ens_acc_abs} respectively. We also include the results for all datasets on the accuracy of non-selective ensembling and their ability to mitigate disagreement, in Table~\ref{app:all_ensemble_no_abst_accs} and Table~\ref{app:all_ensemble_no_abst} respectively. 

\begin{table}
\resizebox{\textwidth}{!}{%
\begin{tabular}{ll|lllllll}
% \multicolumn{4}{c}{portion of Taiwanese Credit test data with $\pflip > 0$} \\
%Randomness& $n$ & Mean & Std \\
\multicolumn{7}{c}{mean $\pm$ std. dev of portion of test data with $\pflip > 0$} \\
Randomness& $n$ & Ger. Credit & Adult  &  Seizure & Tai. Credit & Warfarin  & FMNIST &  Colon\\
\midrule
 %& & Mean & Std.\\
RS & 1 & $.570 \pm .020$ & $.087 \pm .001 $ & $.060 \pm .01$ &   $.082 \pm .002$ & $.098 \pm .003$ & $.061 \pm .008$  & $.037 \pm .005$   \\
RS & (5, 10, 15, 20) & $0.0 \pm 0.0$ & $0.0\pm 0.0$ & $0.0 \pm 0.0$ & $0.0 \pm 0.0$ & $0.0 \pm 0.0$ & $0.0 \pm 0.0$ & $0.0 \pm 0.0$  \\
LOO & 1 &  $.262 \pm .014$ & $.063 \pm .001$ & $.031 \pm .001$ & $.031 \pm .001$ & $.033 \pm .003$ & $.034 \pm .004$ & $.042 \pm .005$ \\
LOO &(5, 10, 15, 20) & $0.0 \pm 0.0$ & $0.0\pm 0.0$ & $0.0 \pm 0.0$ & $0.0 \pm 0.0$ & $0.0 \pm 0.0$ & $0.0 \pm 0.0$ & $0.0 \pm 0.0$ \\
\bottomrule
\end{tabular}
}
\caption{Percentage of points with disagreement between at least one pair of models ($\pflip > 0$) trained with different random seeds (RS) or leave-one-out differences in training data, for singleton models ($n=1$) and selective ensembles ($n > 1$). We present the mean and standard deviation of this percentage over 10 runs of re-sampling ensemble models. Note that these results are for  \textbf{alpha=0.05}, which are presented in the main paper.}
\label{app:tab:ens_disagree_std}
\end{table}

\begin{table}
\resizebox{\textwidth}{!}{%
\begin{tabular}{ll|l@{\hskip 1pt}l@{\hskip 1pt}l@{\hskip 1pt}l@{\hskip 1pt}l@{\hskip 1pt}l@{\hskip 1pt}l@{\hskip 1pt}l}
% \toprule
    &     \multicolumn{8}{c}{\emph{mean accuracy (abstain as error) / std. dev}} \\
\randomness & $n$ & Ger. Credit & Adult &  Seizure & Warfarin & Tai. Credit & FMNIST & Colon \\
\midrule
RS & 5 & $0.0\pm0.0$ & $0.0\pm0.0$ & $0.0\pm0.0$ & $0.0\pm0.0$ & $0.0\pm0.0$ & $0.0\pm0.0$ & $0.0\pm0.0$ \\
RS & 10 & $.576\pm.013$  & $.820\pm8e-4~$ & $.960\pm1e-3$ & $.660\pm2e-3$ &  $.800\pm1e-3$   & $.888\pm2e-3$ & $.914\pm1e-3$ \\
RS & 15 & $.636\pm.017$  & $.827\pm5e-4~$  & $.965\pm1e-3$ &  $.668\pm2e-3$ & $.807\pm9e-4$  & $.897\pm2e-3$ & $.919\pm1e-3$  \\
RS & 20 & $.664\pm.014$  & $.830\pm5e-4~$  & $.967\pm9e-4$ &  $.670\pm3e-3$& $.810\pm8e-4$  & $.902\pm1e-3$   & $.921\pm1e-3$ \\
\midrule
LOO & 5  & $0.0\pm0.0$ & $0.0\pm0.0$ & $0.0\pm0.0$ & $0.0\pm0.0$ & $0.0\pm0.0$ & $0.0\pm0.0$ & $0.0\pm0.0$ \\
LOO & 10  & $.653\pm.017$~~~~~~ & $.827\pm1e-3$~~~~~~ & $.962\pm2e-3$~~~~~~  & $.677\pm1e-3$~~~~~~& $.812\pm1e-3$~~~~~~ & $.909\pm4e-4$~~~~~~  & $.912\pm1e-3$~~~~~~ \\
LOO & 15  & $.678\pm.014$ &$.832\pm7e-4~$ & $.968\pm9e-4$ & $.679\pm9e-4$ & $.814\pm9e-4$  & $.910\pm1e-3$  & $.916\pm2e-3$ \\
LOO & 20  & $.689\pm.014$ & $.834\pm7e-4~$ & $.970\pm1e-3$ & $.680\pm7e-4$& $.815\pm8e-4$ & $.911\pm4e-4$ & $.918\pm8e-4$  \\
% \end{tabular}
 \\
% \begin{tabular}{ll|l@{\hskip 1pt}l@{\hskip 1pt}l@{\hskip 1pt}l@{\hskip 1pt}l@{\hskip 1pt}l@{\hskip 1pt}l@{\hskip 1pt}l}
 \toprule
         & \multicolumn{7}{c}{\emph{mean abstention rate / std dev}} \\
\randomness & $n$ & Ger. Credit & Adult &  Seizure & Warfarin & Tai. Credit & FMNIST & Colon \\
\midrule
RS & 5 & $1.0 \pm 0.0 $ & $1.0 \pm 0.0$ & $1.0 \pm 0.0$ & $1.0 \pm 0.0$ & $1.0 \pm 0.0$ & $1.0 \pm 0.0$ & $1.0 \pm 0.0$ \\
RS & 10 & $.291\pm .014$ & $ .043\pm1e-3$ & $ .02\pm1e-3$  & $.050\pm3e-3$ &  $.039\pm2e-3$ & $.059\pm2e-3$ & $.032\pm3e-3$ \\
RS & 15 & $.205\pm .020$ & $.032\pm1e-3$  & $.018\pm1e-3$ & $.037\pm3e-3$ & $.028\pm1e-3$  & $.042\pm2e-3$ & $.023\pm2e-3$ \\
RS & 20 & $.165\pm.015$ & $.024\pm7-e4$  & $.014\pm7e-4$  &  $.031\pm4e-3$ & $.023\pm8e-4$  & $.036\pm1e-3$  & $.019\pm2e-3$ \\
\midrule
LOO & 5  & $1.0 \pm 0.0 $ & $1.0 \pm 0.0$ & $1.0 \pm 0.0$ & $1.0 \pm 0.0$ & $1.0 \pm 0.0$ & $1.0 \pm 0.0$ & $1.0 \pm 0.0$ \\
LOO & 10  & $.151 \pm .041$~~~~~~ & $.032\pm2e-3$~~~~~~ & $.027\pm2e-3$~~~~~~& $.018\pm2e-3$~~~~~~& $.017\pm2e-3$~~~~~~ & $.020\pm5e-4$~~~~~~   & $.036\pm3e-3$~~~~~~ \\
LOO & 15  & $.105\pm 0.034$  &$.022\pm1e-3$ & $.019\pm1e-3$ &$.013\pm2e-3$& $.013\pm2e-3$ & $.016\pm2e-3$  & $.027\pm2e-3$ \\
LOO & 20  & $.079\pm.029$ &  $.018\pm1e-3$& $.015\pm1e-3$ & $.011\pm2e-3$& $.010\pm1e-3$ & $.012\pm8e-4$ & $.023\pm2e-3$ \\
\bottomrule
\end{tabular}
}
\caption{Accuracy (above) and abstention rate (below) of selective ensembles with $n \in \{5, 10, 15, 20\}$ constituents.
Results are averaged over 24 models, standard deviation is presented. Note that these results are for  \textbf{alpha=0.05}, which are presented in the main paper.}
\label{app:tab:ens_acc_abs}
\end{table}

\begin{figure}[t]
\resizebox{\textwidth}{!}{%
\begin{tabular}{ll | @{\hskip 5pt}c@{\hskip 5pt}c@{\hskip 5pt}c@{\hskip 5pt}c@{\hskip 5pt}c@{\hskip 5pt}c@{\hskip 5pt}c}
% \toprule
         & \multicolumn{8}{c}{\emph{disagreement of non-abstaining ensembles}} \\
\randomness & $n$ & Ger. Credit & Adult  & Seizure & Tai. Credit & Warfarin  & FMNIST & Colon \\
\midrule
RS & 1 & $.570 \pm .020$ & $.087 \pm .001 $ & $.060 \pm .01$ &  $.082 \pm .002$ & $.098 \pm .003$  &  $0.113 \pm .005$  & $.066 \pm .002$   \\
RS & 5 & $.305 \pm .017$ & $.045 \pm .001$ & $.028 \pm .001$ & $.082 \pm .002$ & $.054 \pm .003$  &  $.046\pm .002$ & $.022 \pm .001$ \\
RS & 10 & $.234 \pm .014$ &  $.031 \pm .001$ &  $.019 \pm .001$ &  $.041 \pm .001$ &  $.040 \pm .002$ &  $.032\pm .002$ &  $.014 \pm .002$ \\
RS & 15 & $.185 \pm .012$ &  $.026 \pm .001$ &  $.015 \pm .001$ &   $.030 \pm .000$ & $.033 \pm .002$ &  $.028\pm .002$ &  $.012\pm .001$  \\
RS & 20 & $.155 \pm .010$ &  $.022 \pm .001$ &  $.013 \pm .001 $ &   $.021 \pm .001$ & $.030 \pm .002$ &  $.026\pm .001$ & $.010\pm .001$  \\
\midrule
LOO & 1 &  $.262 \pm .014$ & $.063 \pm .001$ & $.031 \pm .001$ & $.031 \pm .001$ & $.033 \pm .003$ & $.056 \pm .004$ & $.068 \pm .003$ \\
LOO & 5  & $.142 \pm .037$ &  $.033 \pm .001$ & $.028 \pm .001$ & $.019 \pm .001$  & $.018 \pm .001$  & $.032\pm .002$  & $.030 \pm .003$  \\
LOO & 10  & $.111 \pm .020$  & $.023 \pm .001$  & $.020 \pm .001$ & $.014 \pm .001$ & $.016 \pm .001$& $.034\pm .002$  & $.016 \pm .003$  \\
LOO & 15  & $.074 \pm .020$ & $.019 \pm .001$ &  $.017 \pm .001$ &  $.011 \pm .001$ & $.012 \pm .001$ &  $.029\pm .001$ & $.014 \pm .002$ \\
LOO & 20  & $.067 \pm .013$ &  $.016 \pm .001$ & $.015 \pm .001$ &  $.010 \pm .000$ & $.011 \pm .001 $ &  $.027\pm .001$ &  $.010 \pm .001$ \\

\bottomrule
\end{tabular}%ensebmle_no_abst_all}
}
%CHANGE TO JUST A TABLE
\caption{Mean and standard deviation of the percentage of test data with non-zero disagreement over 24 normal (i.e., not selective) ensembles. The mean and standard deviation are taken over ten re-samplings of 24 ensembles.%Horizontal axis depicts ensemble size. 
While ensembling alone mitigates much of the prediction instability, it is unable to eliminate it as selective ensembles do.%The blue line depicts the percentage of points in the validation split where $P_\text{flip}>0$ over changes in random seed, and the green line depicts the same for changes over leave-one-out differences to the training set.
}
\label{app:all_ensemble_no_abst}
\end{figure}

\begin{figure}[t]
\resizebox{\textwidth}{!}{%
\begin{tabular}{ll | @{\hskip 5pt}c@{\hskip 5pt}c@{\hskip 5pt}c@{\hskip 5pt}c@{\hskip 5pt}c@{\hskip 5pt}c@{\hskip 5pt}c}
% \toprule
         & \multicolumn{8}{c}{\emph{accuracy of non-abstaining ensembles}} \\
\randomness & $n$ & Ger. Credit & Adult  & Seizure & Warfarin  & Tai. Credit & FMNIST & Colon \\
\midrule
RS & 5 &$0.745 \pm 0.013$ & $0.842 \pm 0.001$ & $0.975 \pm 0.001$ & $0.688 \pm 0.0$ & $0.822 \pm 0.001$ & $0.919 \pm 0.001$ & $0.927 \pm 0.001$ \\
RS & 10	& $0.747 \pm 0.014$ & $0.843 \pm 0.001$ & $0.975 \pm 0.001$ & $0.688 \pm 0.0$ & $0.822 \pm 0.001$ & $0.92 \pm 0.001$ & $0.928 \pm 0.001$ \\
RS & 15 & $0.75 \pm 0.01$ & $0.842 \pm 0.001$ & $0.975 \pm 0.001$ & $0.688 \pm 0.0$ & $0.822 \pm 0.001$ & $0.92 \pm 0.001$ & $0.928 \pm 0.001$ \\
RS & 20 & $0.747 \pm 0.01$ & $0.842 \pm 0.0$ & $0.975 \pm 0.001$ & $0.688 \pm 0.0$ & $0.822 \pm 0.001$ & $0.92 \pm 0.001$ & $0.928 \pm 0.0$ \\
\midrule
LOO & 5 & $0.728 \pm 0.011$ & $0.844 \pm 0.0$ & $0.979 \pm 0.001$ & $0.685 \pm 0.002$ & $0.821 \pm 0.001$ & $0.918 \pm 0.0$ & $0.927 \pm 0.002$ \\
LOO & 10 &$0.728 \pm 0.008$ & $0.844 \pm 0.001$ & $0.978 \pm 0.001$ & $0.686 \pm 0.002$ & $0.821 \pm 0.001$ & $0.918 \pm 0.0$ & $0.927 \pm 0.002$ \\
LOO & 15 &$0.733 \pm 0.008$ & $0.844 \pm 0.0$ & $0.979 \pm 0.001$ & $0.685 \pm 0.001$ & $0.821 \pm 0.0$ & $0.917 \pm 0.0$ & $0.927 \pm 0.001$ \\
LOO & 20 & $0.73 \pm 0.008$ & $0.843 \pm 0.0$ & $0.979 \pm 0.001$ & $0.685 \pm 0.002$ & $0.821 \pm 0.0$ & $0.918 \pm 0.001$ & $0.927 \pm 0.001$ \\
\bottomrule
\end{tabular}%ensebmle_no_abst_all}
}
%CHANGE TO JUST A TABLE
\caption{Accuracy of non-selective (regular) ensembles with $n \in \{5, 10, 15, 20\}$ constituents.
Results are averaged over 24 models, standard deviation is presented. 
}
\label{app:all_ensemble_no_abst_accs}
\end{figure}

\section{Selective Ensembles and Disparity in Selective Prediction}
\label{appendix:fairness}
In light of the fact that prior work has brought to light the possibility of selective prediction exacerbating model accuracy disparity between demographic groups~\cite{jones2020selective}, we present the selective ensemble accuracy and abstention rate group-by-group for several different demographic groups across four datasets: Adult, German Credit, Taiwanese Credit, and Warfarin Dosing. Results are in Table~\ref{app:tab:fairness}.

 \begin{table}
\resizebox{\textwidth}{!}{%
\begin{tabular}{ll|r@{\hskip 0pt}lr@{\hskip 0pt}lr@{\hskip 0pt}lr@{\hskip 0pt}lr@{\hskip 0pt}lr@{\hskip 0pt}lr@{\hskip 0pt}lr@{\hskip 0pt}lr@{\hskip 0pt}l}
\toprule
         & \multicolumn{18}{c}{\emph{accuracy (abstain as error) / abstention rate}} \\
\randomness & $n$ & \multicolumn{2}{c}{Adult Male} &  \multicolumn{2}{c}{Adult Fem. }&  \multicolumn{2}{c}{Ger. Cred. Young} & \multicolumn{2}{c}{Ger. Cred. Old} & \multicolumn{2}{c}{Tai. Cred. Male} & \multicolumn{2}{c}{Tai. Cred. Fem.} & \multicolumn{2}{c}{Warf. Black}  & \multicolumn{2}{c}{Warf. White}  & \multicolumn{2}{c}{Warf. Asian}  \\
\midrule
Base & 1 & $.804$ & / - & $.923$ & / - & $.677$ & / - & $.777$ & / - &  $.814$ & / - & $.825$ & / - & $.665$ & / - & $.688$ & / - & $.689$ & / - \\ 
RS & 5 & $0.0$ & $/ 1.0$ & $0.0$ & $/ 1.0$ & $0.0$ & $/ 1.0$ & $0.0$ & $/ 1.0$ & $0.0$ & $/ 1.0$  & $0.0$ & $/ 1.0$ & $0.0$ & $/ 1.0$ & $0.0$ & $/ 1.0$ & $0.0$ & $/ 1.0$  \\
RS & 10 & $.777$ & $/ .053$ & $.912$ & $/ .023$ & $.507$ & $/ .334$  & $.636$ & $/ .254$ &  $.791$ & $/ .048$  & $.807$ & $/ .035$ & $.659$ & $/ .009$ & $.681$ & $/ .002$ & $.683$ & $/ .007$\\
RS & 15 & $.786$ & $/ .037$ & $.915$ & $/ .015$  & $.559$ & $/ .248$ &  $.705$ & $/ .168$ & $.798$ & $/ .033$  & $.812$ & $/ .025$ & $.664$ & $/ .010$ &$.683$ & $/ .002$& $.688$ & $/ .006$\\
RS & 20 & $.789$ & $/ .030$ & $.917$ & $/ .013$  & $.586$ & $/ .205$  &  $.733$ & $/ .130$ & $.802$ & $/ .028$  & $.814$ & $/ .020$  & $.667$ & $/ .009$ &$.683$ & $/ .002$ & $.689$ & $/ .006$\\
\midrule
Base & 1 & $.806$ & / - & $.922$ & / - & $.697$ & / - & $.757$ & / - & $.815$ & / - & $.825$ & / - & $.665$ & / - & $.687$ & / - & $.688$ & / - \\
LOO & 5  & $0.0$ & $/ 1.0$ & $0.0$ & $/ 1.0$ & $0.0$ & $/ 1.0$  & $0.0$ & $/ 1.0$ & $0.0$ & $/ 1.0$ & $0.0$ & $/ 1.0$ & $0.0$ & $/ 1.0$ & $0.0$ & $/ 1.0$ & $0.0$ & $/ 1.0$\\
LOO & 10  & $.787$ & $/ .038$ & $.913$ & $/ .018$ & $.612$ & $/ .166$ & $.689$ & $/ .138$& $.802$ & $/ .023$ & $.817$ & $/ .014$   & $.655$ & $/ .020$ & $.680$ & $/ .019$ & $.680$ & $/ .019$ \\
LOO & 15  & $.793$ & $/ .026$  &$.916$ & $/ .012$ & $.646$ & $/ .101$ & $.704$ & $/ .107$& $.806$ & $/ .017$ & $.819$ & $/ .011$  & $.658$ & $/ .014$ & $.681$ & $/ .013$ & $.682$ & $/ .013$\\
LOO & 20  & $.796$ & $/ .022$  & $.917$ & $/ .010$& $.661$ & $/ .071$ & $.714$ & $/ .084$& $.808$ & $/ .014$ & $.820$ & $/ .009$ & $.659$ & $/ .011$ & $.682$ & $/ .011$ & $.683$ & $/ .011$\\
\bottomrule
\end{tabular}
}
\caption{We present the selective ensemble accuracy and abstention rate group-by-group for several different demographic groups across four datasets: Adult, German Credit, Taiwanese Credit, and Warfarin Dosing. We note that by and large, using selective ensembles did not exacerbate accuracy disparity by very much (within ~1\% of the original disparity), although they did not ameliorate disparities in accuracy that already existed within the performance of the algorithm. The only exception to this was German Credit, where we note, as in the remainder of our results, that the entire dataset is only ~1000 points, so results may be slightly different in this regime. Overall, we note that subgroup abstention rates can vary by dataset, and so it should be studied whenever selective ensembles are used in a sensitive setting.}
\label{app:tab:fairness}
\end{table}

\section{Explanation Consistency Full Results}
\label{appendix:explanation_stability}
We give full results for selective and non-selective ensembling's mitigation of inconsistency in feature attributions. 
\subsection{Attributions}
\label{appendix:attributions}
\begin{figure}[t]
\resizebox{\textwidth}{!}{%
\begin{subfigure}{\textwidth}
\resizebox{\textwidth}{!}{%
\includegraphics{pics/test_gc_pic1.pdf}
\includegraphics{pics/test_gc_pic2.pdf}
}
\\
\caption{}
\label{gc_rs}
\end{subfigure}%
}
\resizebox{\textwidth}{!}{%
\begin{subfigure}{\textwidth}
\resizebox{\textwidth}{!}{%
\includegraphics{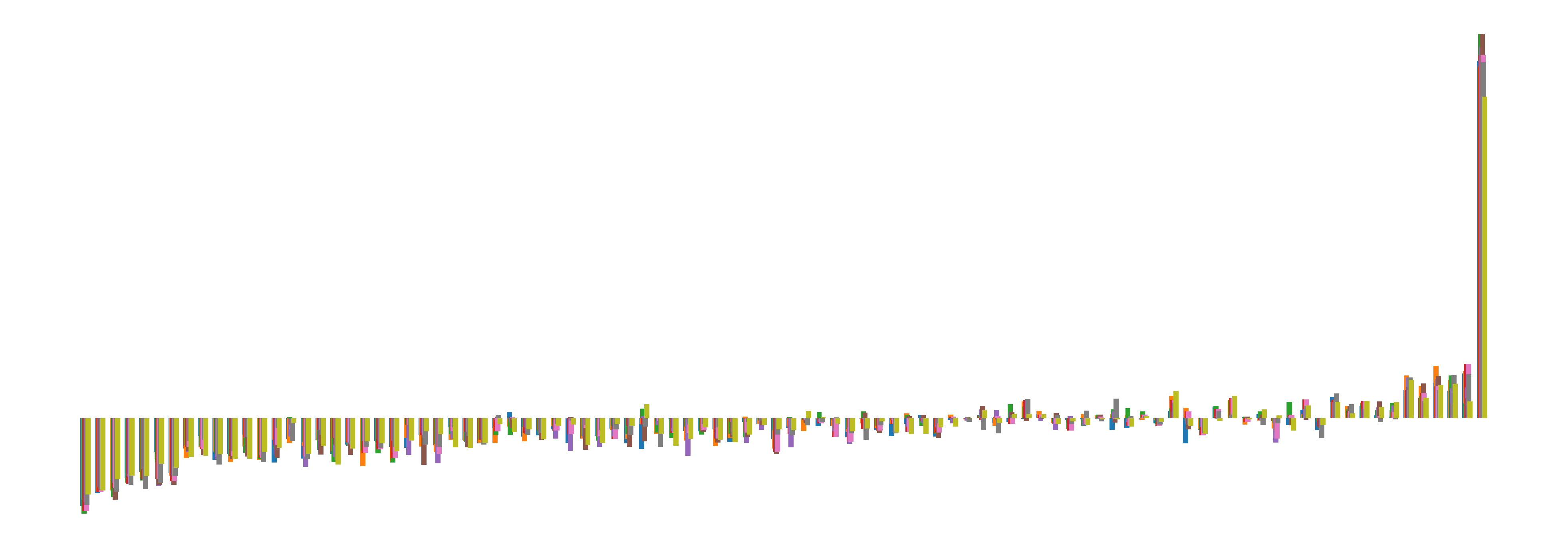}
\includegraphics{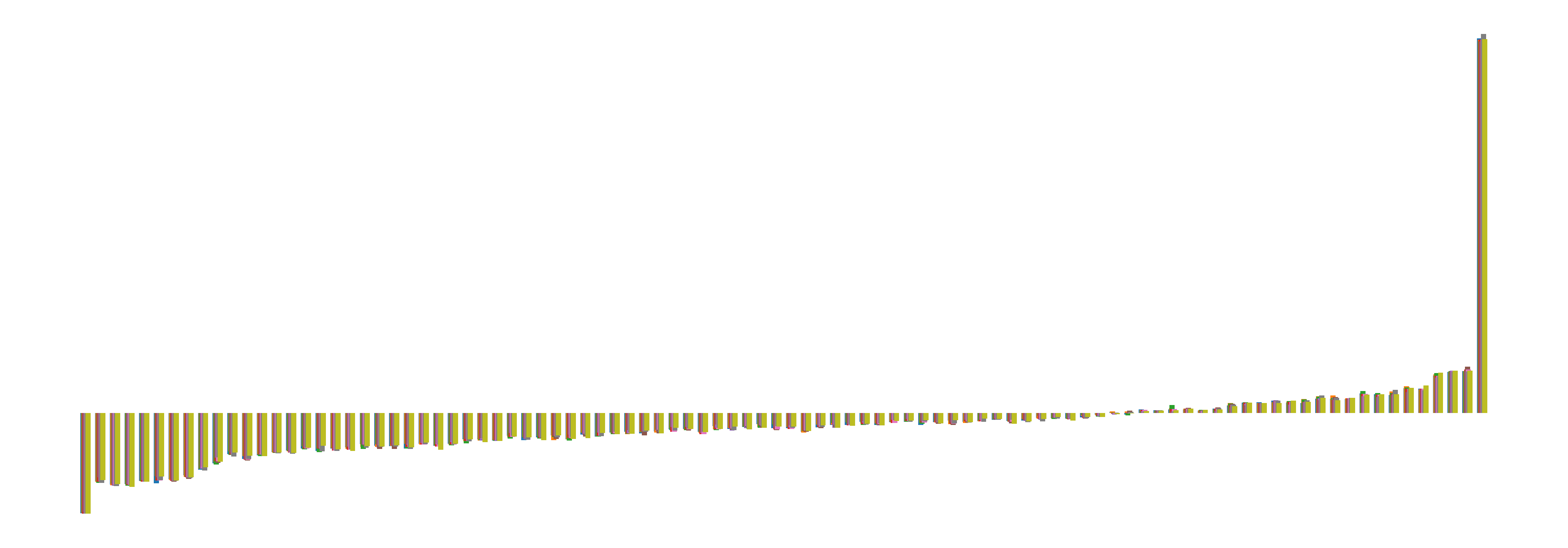}
}
\caption{}
\label{ad_rs}
\end{subfigure}
}
\resizebox{\textwidth}{!}{%
\begin{subfigure}{\textwidth}
\resizebox{\textwidth}{!}{%
\includegraphics{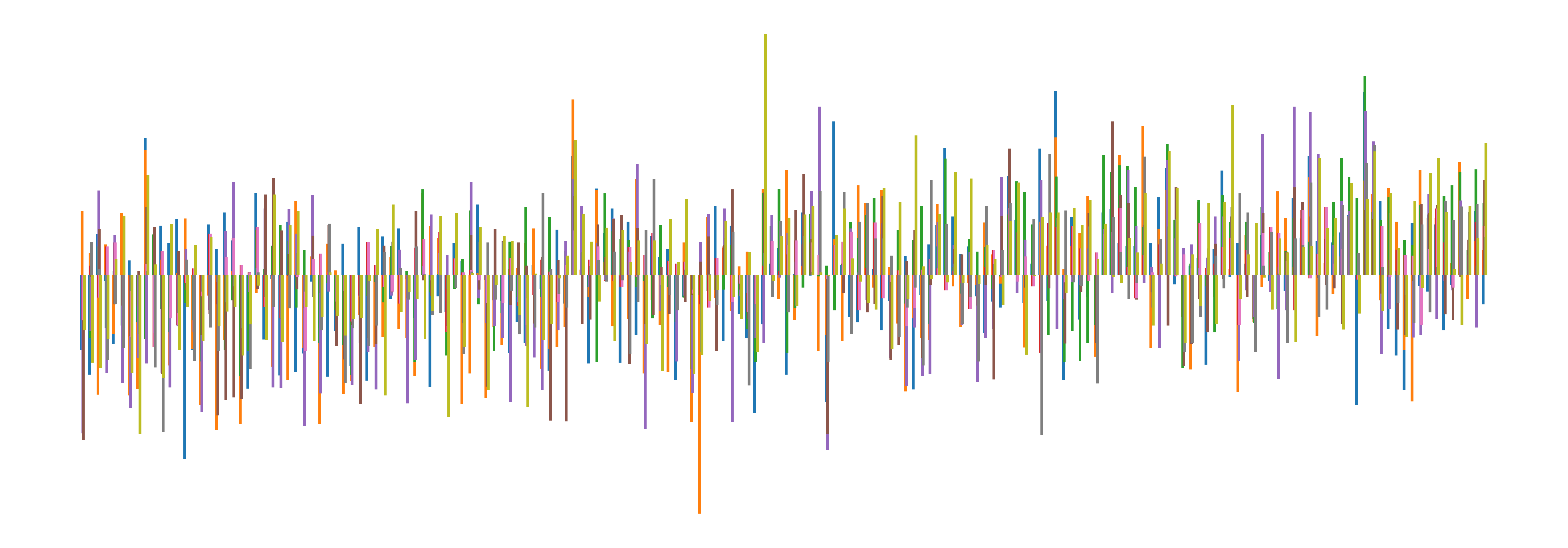}
\includegraphics{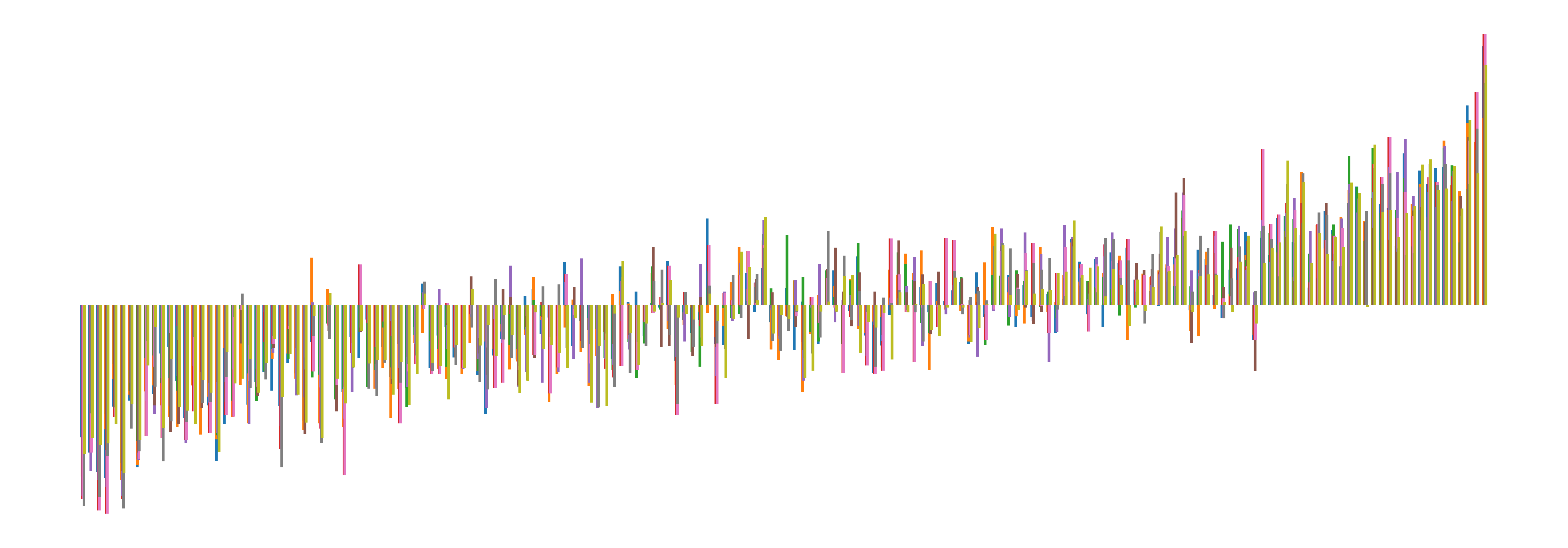}
}
\caption{}
\label{sz_rs}
\end{subfigure}
}
\resizebox{\textwidth}{!}{%
\begin{subfigure}{\textwidth}
\resizebox{\textwidth}{!}{%
\includegraphics{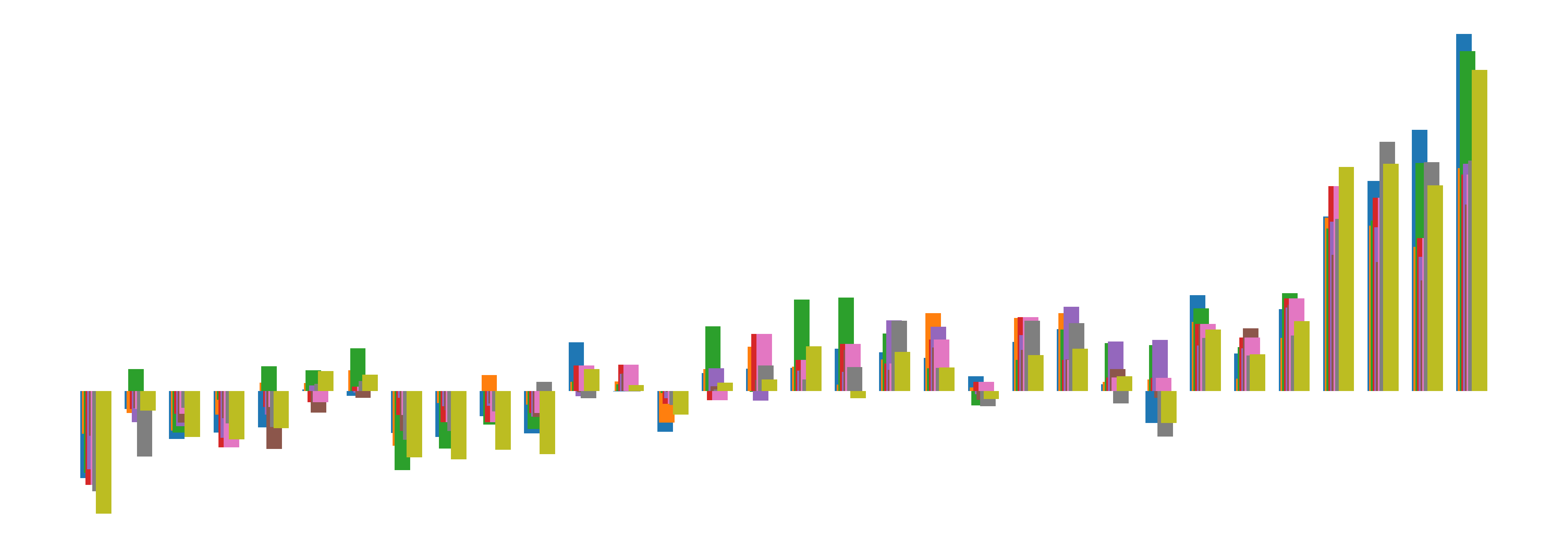}
\includegraphics{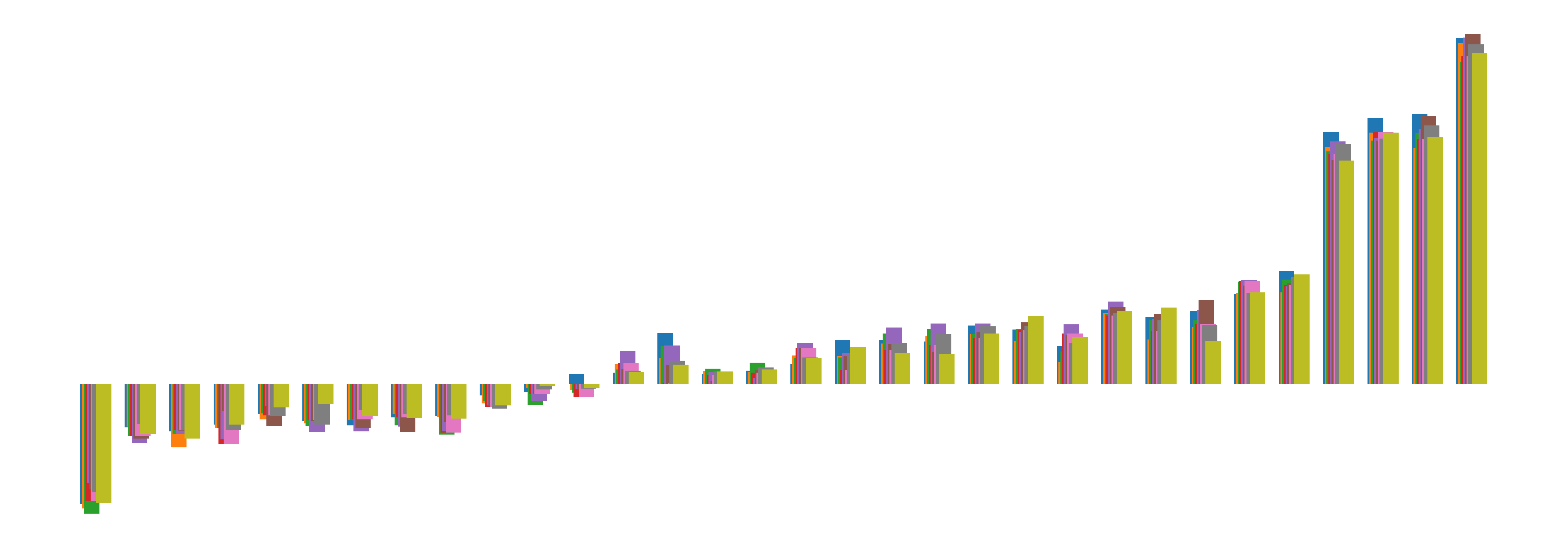}
}
\caption{}
\label{tai_rs}
\end{subfigure}
}
\resizebox{\textwidth}{!}{%
\begin{subfigure}{\textwidth}
\resizebox{\textwidth}{!}{%
\includegraphics{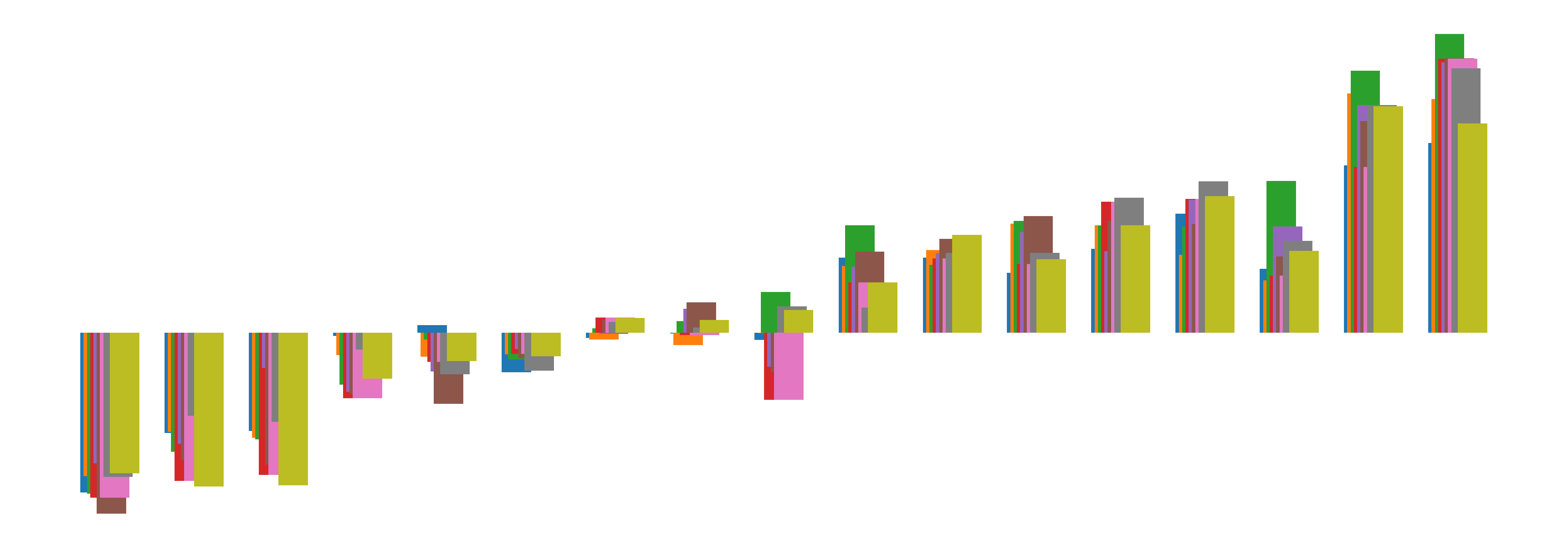}
\includegraphics{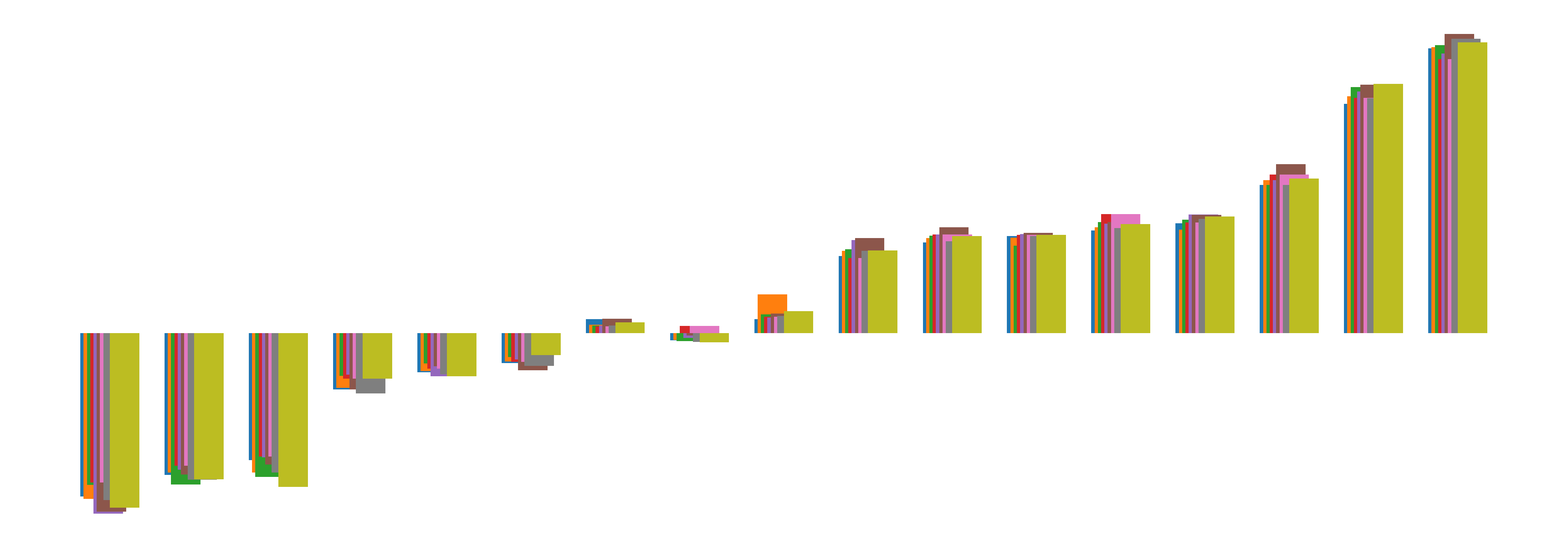}
}
\caption{}
\label{war_rs}
\end{subfigure}
}
 \caption{Inconsistency of attributions on the same point across an individual (left) and ensembled (right) model ($n=15$), for all datasets, over  differences in random seed chosen for initialization parameters before training.
    The height of each bar on the horizontal axis represents the attribution score of a distinct feature, and each color represents a different model.
    Features are ordered according to the attribution scores of one randomly-selected model. Figure~\subref{gc_rs} depicts the German Credit Dataset, Figure~\subref{ad_rs} depicts Adult, Figure~\subref{sz_rs} Seizure, Figure~\subref{tai_rs} Taiwanese, and Figure~\subref{war_rs} Warfarin. We do not include feature attribution for image datasets as the individual pixels are less meaningful than the feature attributions in a tabular dataset.}
\label{app:rs_attrs_pic}
\end{figure}

\begin{figure}[t]
\resizebox{\textwidth}{!}{%
\begin{subfigure}{\textwidth}
\resizebox{\textwidth}{!}{%
\includegraphics{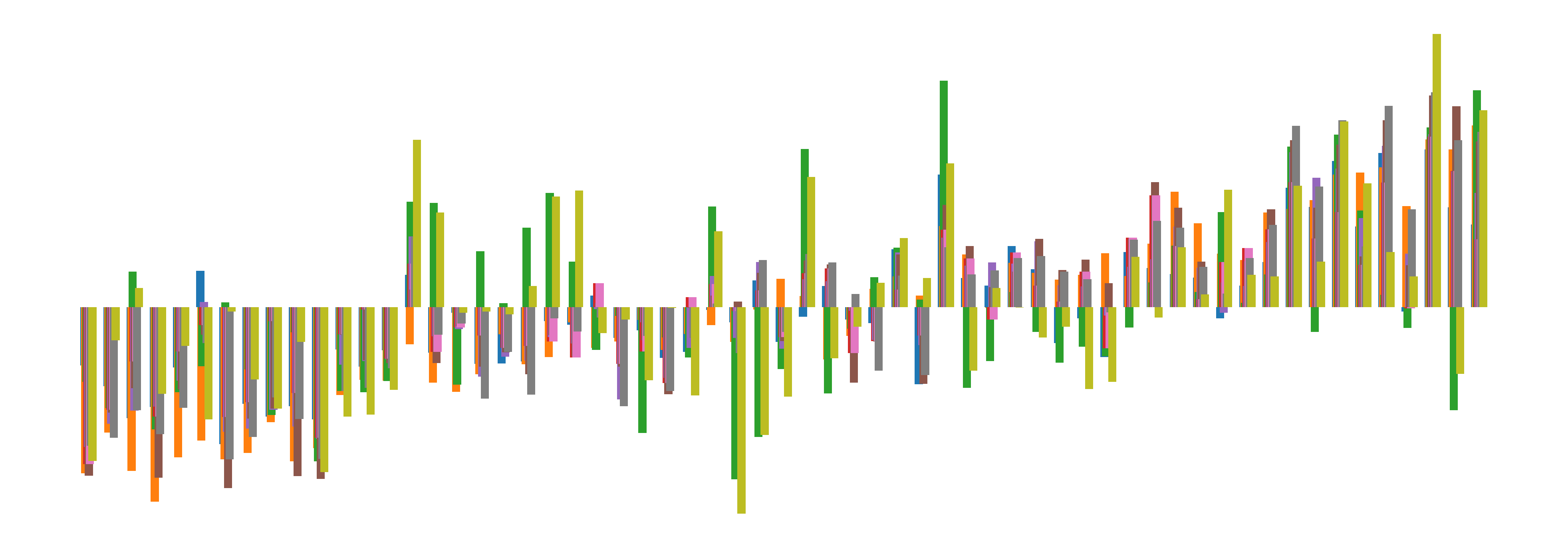}
\includegraphics{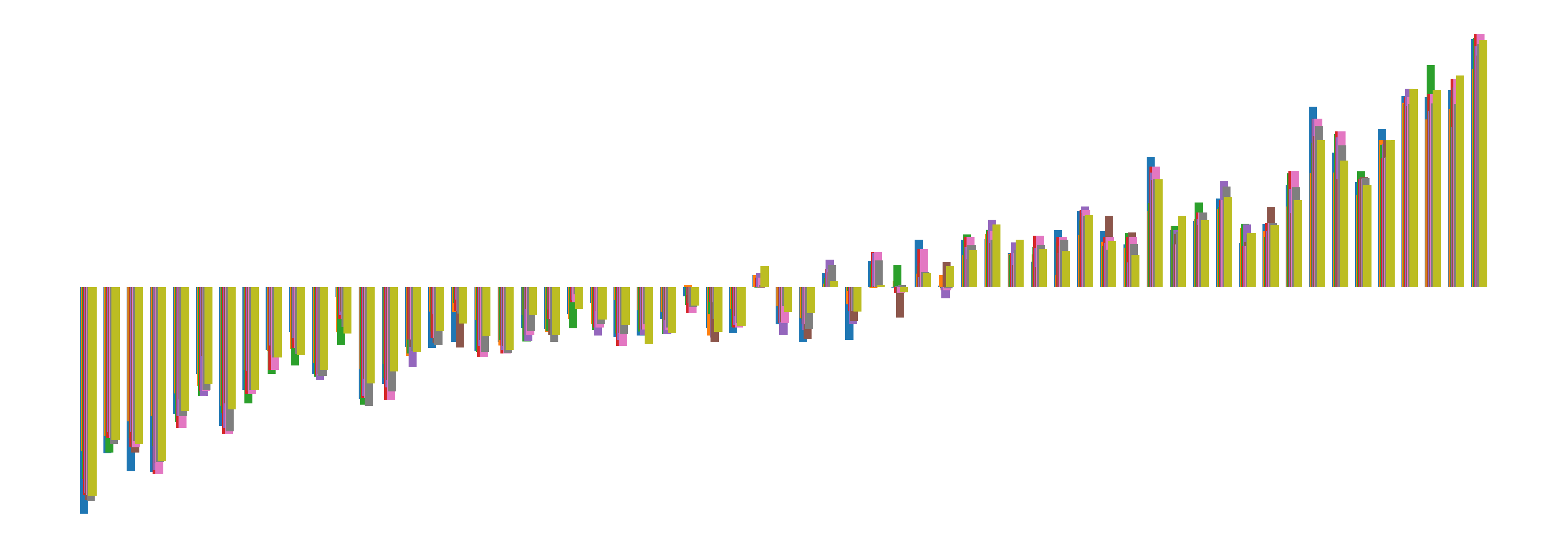}
}
\\
\caption{}
\label{gc_lf}
\end{subfigure}%
}
\resizebox{\textwidth}{!}{%
\begin{subfigure}{\textwidth}
\resizebox{\textwidth}{!}{%
\includegraphics{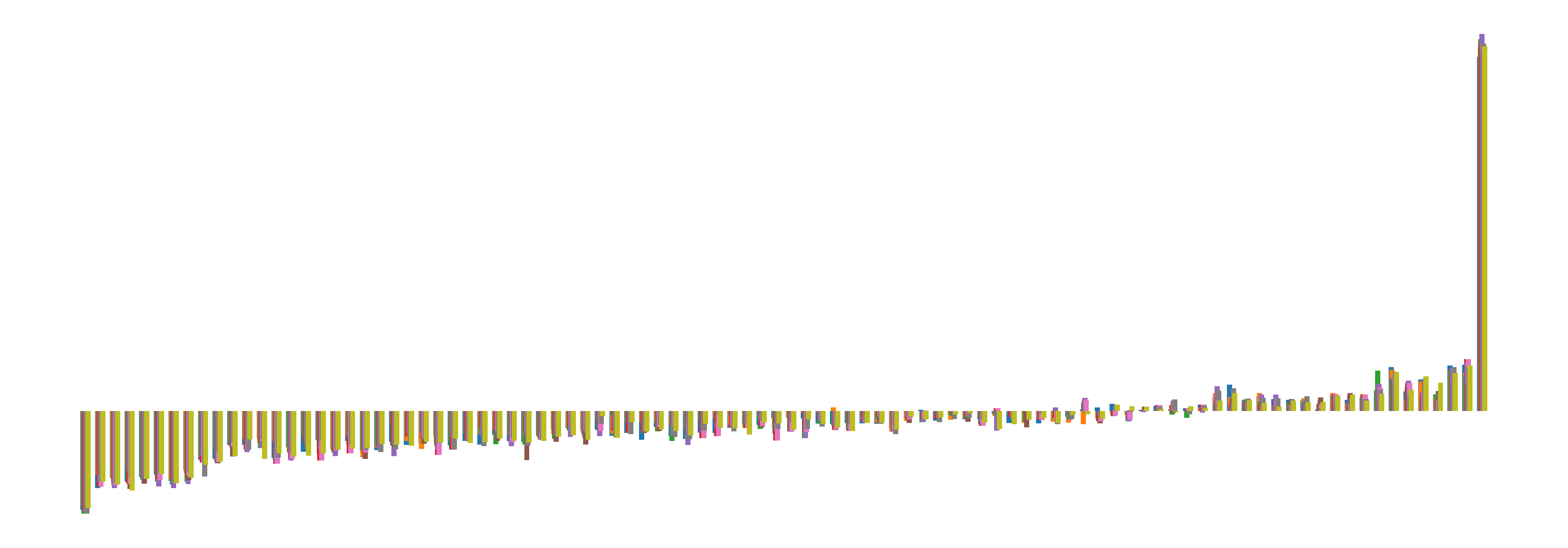}
\includegraphics{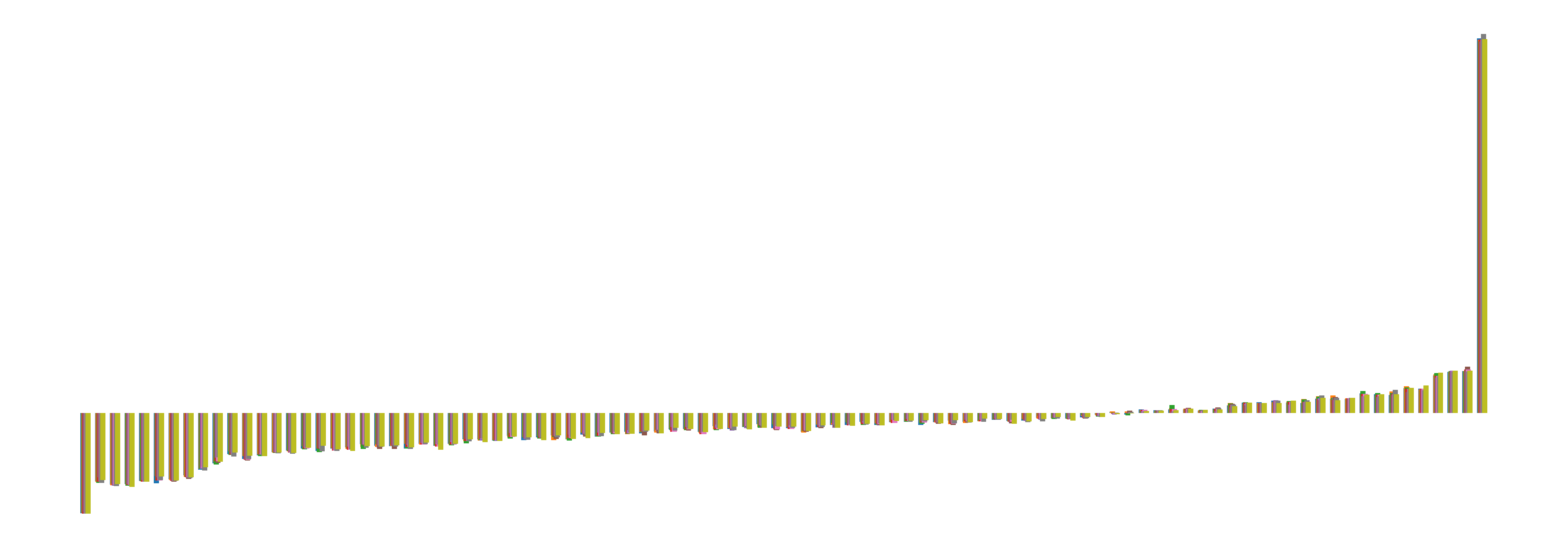}
}
\caption{}
\label{ad_lf}
\end{subfigure}
}
\resizebox{\textwidth}{!}{%
\begin{subfigure}{\textwidth}
\resizebox{\textwidth}{!}{%
\includegraphics{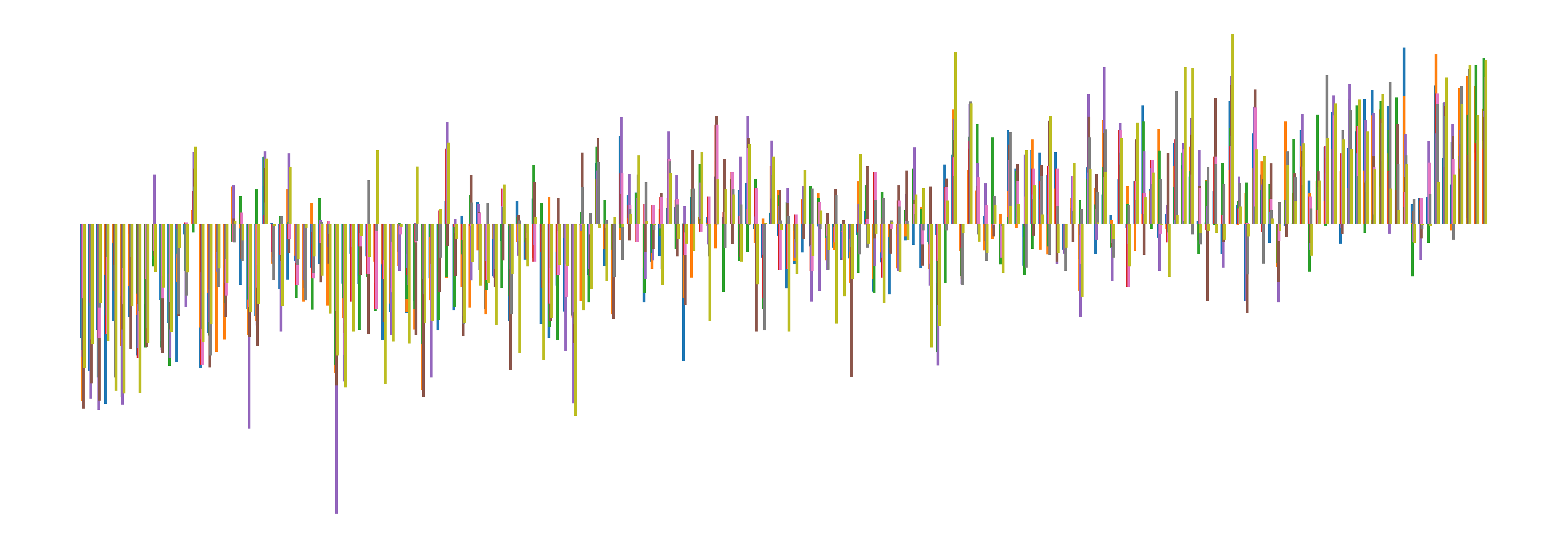}
\includegraphics{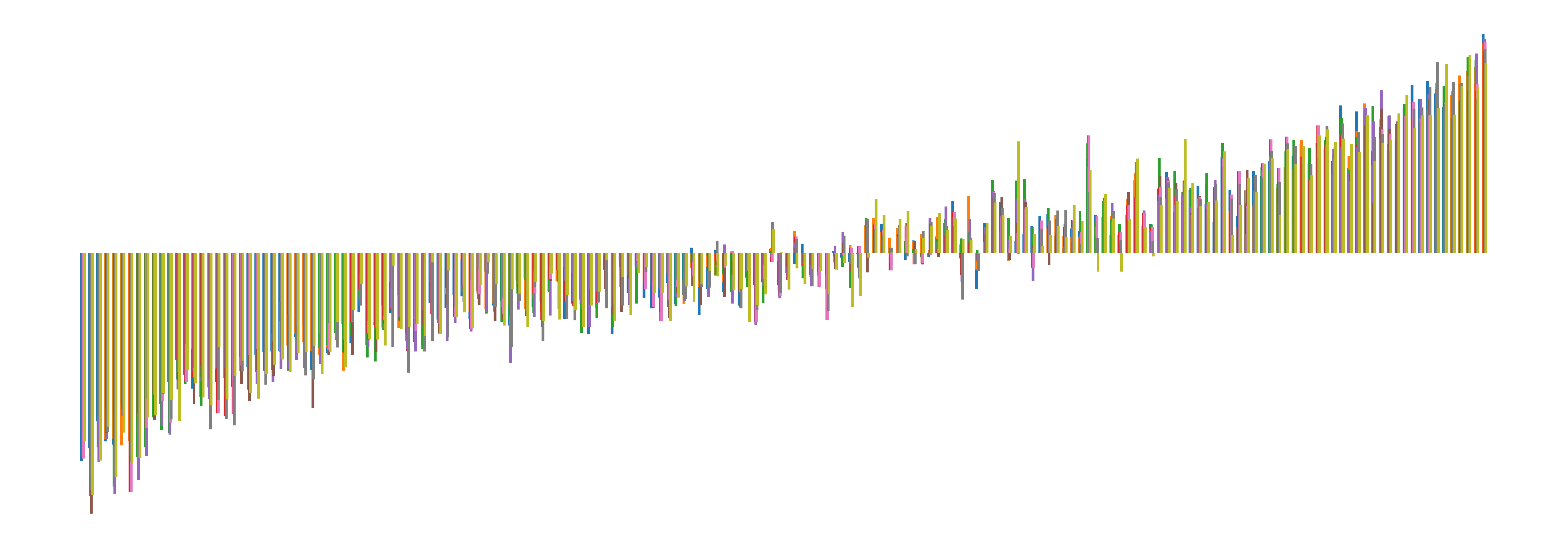}
}
\caption{}
\label{sz_lf}
\end{subfigure}
}
\resizebox{\textwidth}{!}{%
\begin{subfigure}{\textwidth}
\resizebox{\textwidth}{!}{%
\includegraphics{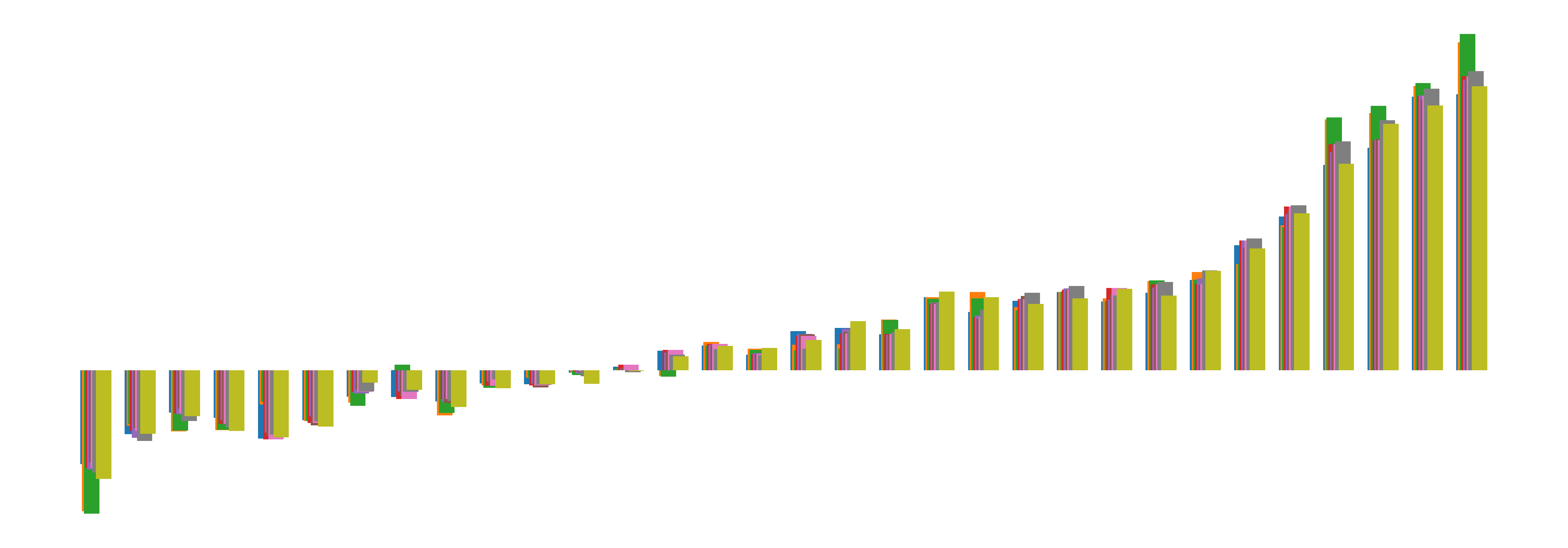}
\includegraphics{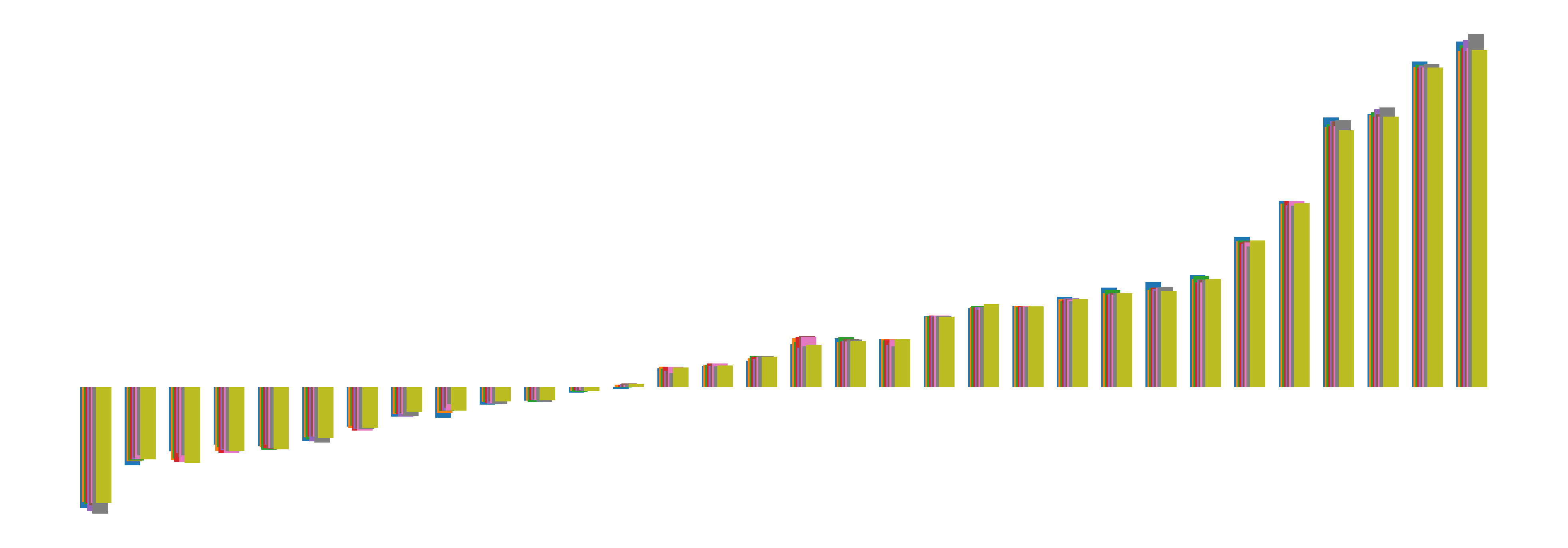}
}
\caption{}
\label{tai_lf}
\end{subfigure}
}
\resizebox{\textwidth}{!}{%
\begin{subfigure}{\textwidth}
\resizebox{\textwidth}{!}{%
\includegraphics{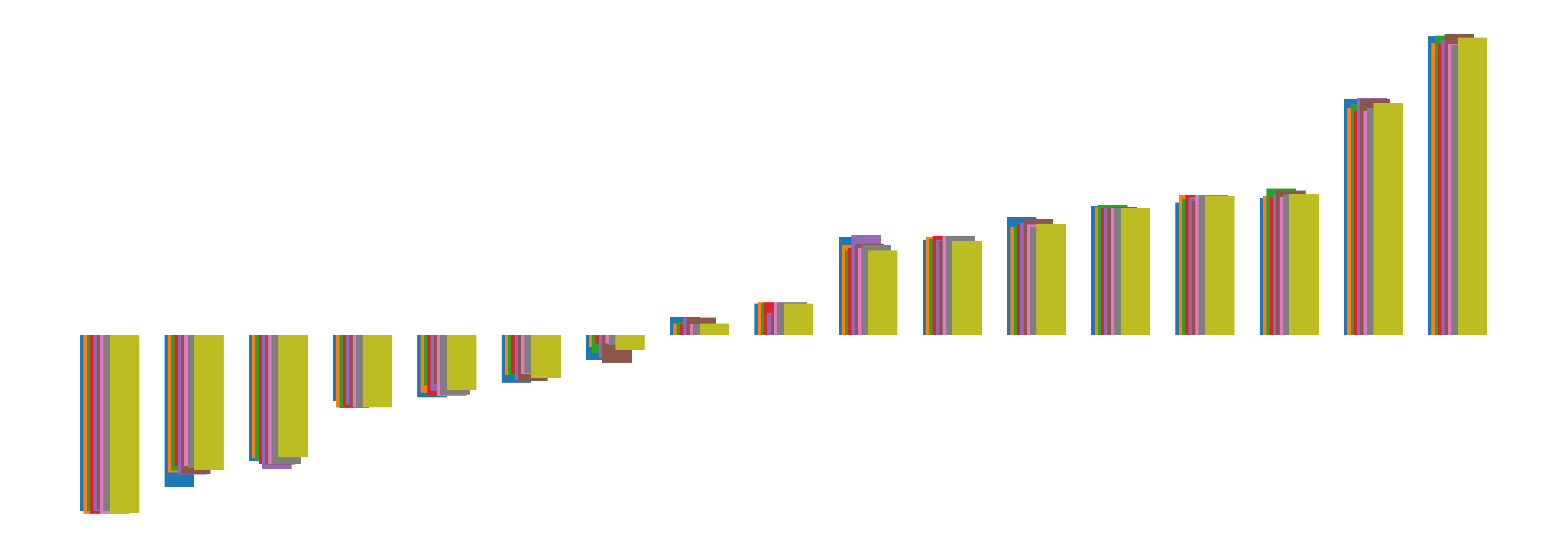}
\includegraphics{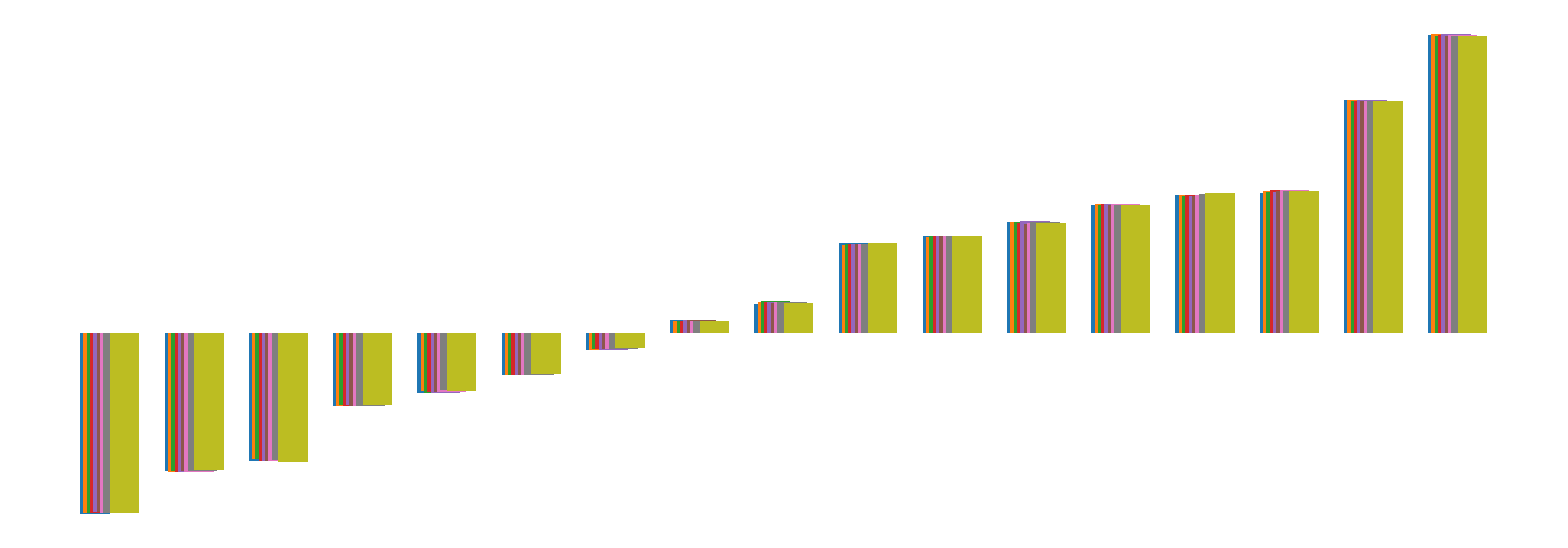}
}
\caption{}
\label{war_lf}
\end{subfigure}
}
 \caption{Inconsistency of attributions on the same point across an individual (left) and ensembled (right) model ($n=15$), for all datasets, over leave-one-out differences in the training set.
    The height of each bar on the horizontal axis represents the attribution score of a distinct feature, and each color represents a different model.
    Features are ordered according to the attribution scores of one randomly-selected model. Figure~\subref{gc_lf} depicts the German Credit Dataset, Figure~\subref{ad_lf} depicts Adult, Figure~\subref{sz_lf} Seizure, Figure~\subref{tai_lf} Taiwanese, and Figure~\subref{war_lf} Warfarin. We do not include feature attribution for image datasets as the individual pixels are less meaningful than the feature attributions in a tabular dataset.}
\label{app:lf_attrs_pic}
\end{figure}

We pictorially show the inconsistency of individual model feature attributions versus the consistency of attributions ensembles of 15 for each tabular dataset in Figure~\ref{app:rs_attrs_pic} and Figure~\ref{app:lf_attrs_pic}. The former shows inconsistency over differences in random initialization, the latter shows inconsistency over one-point changes to the training set.

\subsection{Similarity Metrics of Attributions}
We display how Spearman's ranking coefficient ($\rho$), Pearson's Correlation Coefficient ($r$), top-5 intersection and $\ell_2$ distance between feature attributions \emph{over the same point} become more and more similar with increasing numbers of ensemble models. While the comparisons to generate the similarity score is between two models on the same point, the result is averaged over this comparison for the entire test set. We average this over 276 comparisons between different models. In cases were abstention is high, indicating inconsistency on the dataset for the training pipeline, selective ensembling can further improve stability of attributions by not considering unstable points (see e.g. German Credit). We present the expanded results from the main paper, for all datasets, on all four metrics (as SSIM is only computed for image datasets, and $\rho$ is not computed for image datasets). We display error bars indicating standard deviation over the 276 comparisons between two models for tabular datasets, and 40 comparisons for image datasets.

\begin{figure}[t]
\centering
\resizebox{0.65\textwidth}{!}{%
\includegraphics{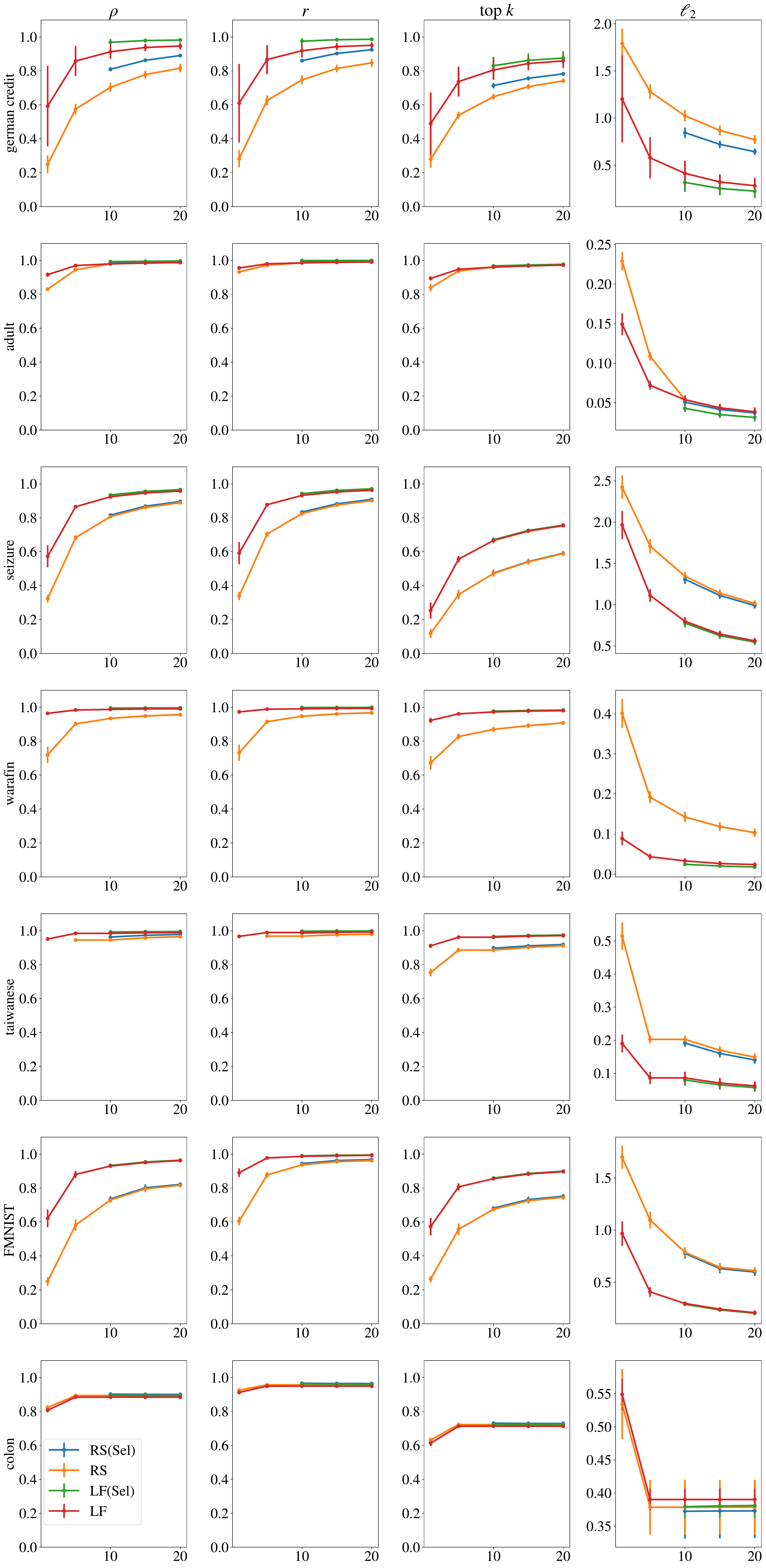}
}
\label{app:all_ensemble_expl_const_all}
\caption{We plot the average similarity across feature attributions for an individual point, averaged over 276 comparisons of feature attributions from two different models. This is aggregated across the entire validation split. The error bars represent the standard deviation over the 276 comparisons between models. Each row of plots constitutes the plots for a given dataset, noted on the far left, and each column of plots is for a given metric, noted at the top. Note that for image datasets, (FMNIST and Colon), we plot SSIM instead of Spearman's Ranking Coefficient $(\rho)$. The x-axis is the number of models in the ensemble, starting with one, and the y-axis indicates the value of the similarity metric averaged over all 276 comparisons of individual points' in the validation split's attributions. The red and orange lines depict regular ensembles, and the green and blue represent selective ensembles.}
\end{figure}

\clearpage

% \bibliographystyle{plainnat}
% \bibliography{bib}

\label{appendix}

\newpage
\end{document}